\documentclass[letterpaper, 10 pt, journal, twoside]{IEEEtran}

\IEEEoverridecommandlockouts                              

\makeatletter
\let\NAT@parse\undefined
\makeatother
\usepackage[square,numbers,sort&compress]{natbib}

\usepackage[pdftex]{graphicx}
\usepackage{amsmath}
\usepackage{amssymb}  
\usepackage{subfig}
\usepackage{multirow}
\usepackage{array,booktabs}
\usepackage{diagbox}
\usepackage{balance}
\usepackage[ruled]{algorithm}
\usepackage{algpseudocode}
\usepackage{colortbl}

\usepackage{mathabx}

\usepackage{fontawesome5}
\usepackage[weather]{ifsym}

 \usepackage[final]{hyperref}
 \hypersetup{
     colorlinks=true,
     linkcolor=blue,
     filecolor=magenta,
     urlcolor=blue,
     citecolor=black
 }
\usepackage[font=small]{caption}
\usepackage{./rpm_packages/rpm_SIunits}
\usepackage{./rpm_packages/rpm_acronyms}
\usepackage{./rpm_packages/rpm_math}
\usepackage{./rpm_packages/rpm_misc}

\usepackage{soul,color}
\usepackage{lipsum}
\usepackage{siunitx}

\usepackage{xcolor}

\usepackage{tikz} 

\title{SHeRLoc: Synchronized Heterogeneous Radar Place Recognition for Cross-Modal Localization}

\author{Hanjun Kim$^{1}$, Minwoo Jung$^{2}$, Wooseong Yang${}^{2}$ and Ayoung Kim${}^{2*}$
\thanks{Manuscript received: June 16, 2025; Revised: September 12, 2025; Accepted:  October 8, 2025. This letter was recommended for publication by Associate Editor S. Behnke and Editor T. Asfour upon evaluation of the reviewers’ comments.
This work was supported in part by the Technology Innovation Program (1415187329, 20024355, Development of autonomous driving connectivity technology based on sensor-infrastructure cooperation) funded by MOTIE, Korea, and in part by the National Research Foundation of Korea (NRF) grant funded by the Korea government (MSIT) (No. RS-2023-00241758).}

\thanks{$^{1}$H. Kim is with the Dept. of Future Automotive Mobility, SNU, Seoul, S. Korea {\tt\small hanjun815@snu.ac.kr}}%
\thanks{$^{2}$M. Jung, W. Yang and A. Kim are with the Dept. of Mechanical Engineering, SNU, Seoul, S. Korea {\tt\small [moonshot, yellowish, ayoungk]@snu.ac.kr}}
\thanks{Project Page: https://sites.google.com/view/radar-sherloc}
\thanks{Digital Object Identifier (DOI): see top of this page.}%
}

\begin{document}

\maketitle
\begin{abstract}

Despite the growing adoption of radar in robotics, the majority of research has been confined to homogeneous sensors, overlooking the integration and cross-modality challenges inherent in heterogeneous radar.
This leads to significant difficulties in generalizing across diverse radar types, with modality-aware approaches that could leverage the complementary strengths of heterogeneous radar remaining unexplored. To bridge these gaps, we propose SHeRLoc, the first deep network tailored for heterogeneous radar, which utilizes radar cross-section polar matching to align multimodal radar data.
Our hierarchical optimal transport-based feature aggregation generates rotationally robust multi-scale descriptors. By employing FFT-similarity-based data mining and adaptive margin-based triplet loss, SHeRLoc enables FOV-aware metric learning. 
SHeRLoc achieves an order of magnitude improvement in heterogeneous radar place recognition, increasing recall@1 from below 0.1 to 0.9 on a public dataset and outperforming state-of-the-art
methods. Also applicable to LiDAR, SHeRLoc paves the way for cross-modal place recognition and heterogeneous sensor SLAM.
The supplementary materials and source code are available at \hyperlink{https://sites.google.com/view/radar-sherloc}{https://sites.google.com/view/radar-sherloc}.


\end{abstract}

\section{Introduction}
\label{sec:intro}

\IEEEPARstart{P}{lace} recognition (PR) is a cornerstone of robust localization in autonomous driving, enabling vehicles to identify previously visited locations. Traditionally, cameras and LiDAR have dominated PR due to their rich data representations~\cite{arandjelovic2016netvlad, kim2018scan}. 
However, their susceptibility to adverse conditions has shifted attention to radar, which offers unparalleled robustness by penetrating fog, rain, and snow~\cite{barnes2020oxford, kim2020mulran}.
Early radar-based PR mostly leveraged $360^\circ$ spinning radars, establishing a foundation for robust localization~\cite{barnes2020under, suaftescu2020kidnapped, jang2023raplace, gadd2024open, kim2024referee}. 
Building on this foundation, compact \ac{SoC} radars have been integrated into PR tasks~\cite{cai2022autoplace, meng2024mmplace, herraez2024spr}.
More recently, 4D radars, capable of capturing range, azimuth, elevation, and radial velocity, have also been adopted in radar PR~\cite{peng2024transloc4d, hilger2025introspective}.

Despite the growing adoption of radar in robotics, the majority of research has been confined to homogeneous sensor systems, overlooking the integration and cross-modality challenges inherent in heterogeneous radar technologies. Notably, different radar systems operate under distinct sensing principles, leading to substantial variations in noise characteristics, point density, and \ac{FOV}.
As a result, methods designed for homogeneous radar systems fail to generalize across diverse radar configurations.
Moreover, radar types are suited for different tasks: spinning radar, with its dense $360^\circ$ measurements, is ideal for building comprehensive mapping databases, while 4D radar, prevalent in modern vehicles, excels in real-time dynamic sensing and is well-suited for query generation.
Despite these complementary properties, existing approaches do not account for the differences between radar modalities, leaving the problem of heterogeneous radar PR largely unresolved.
These challenges highlight the need for a new approach: \textit{modality-aware heterogeneous radar PR}.
\begin{figure}[!t]
    \centering
    \includegraphics[trim= 18.9cm 18.5cm 14.8cm 17.1cm, clip,width=1.0\columnwidth]{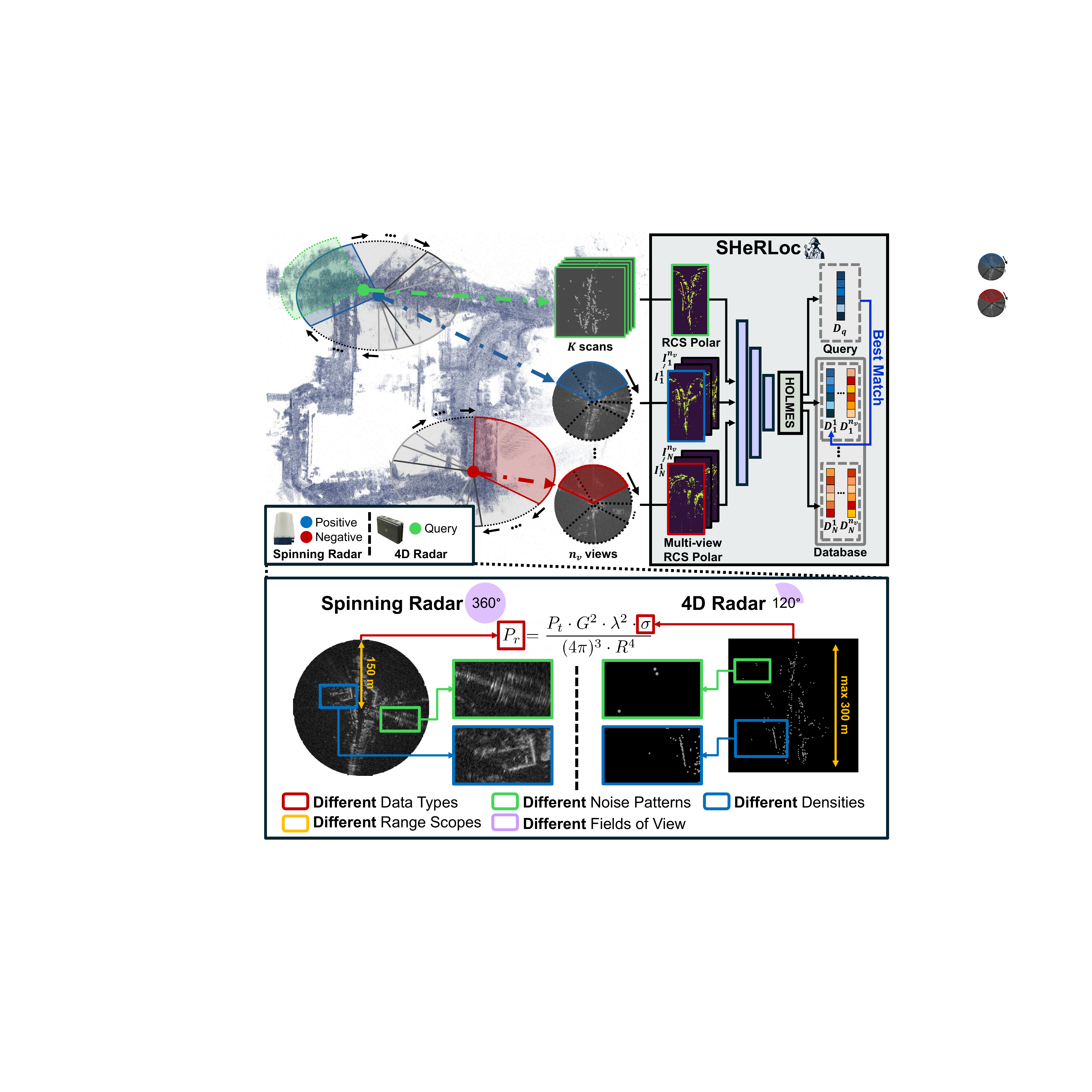}
    \caption{SHeRLoc generates $n_v$ views from spinning radar scans to align with the narrow \ac{FOV} of 4D radar, while bridging the modality gap across heterogeneous radars through RCS polar matching.}
    \label{fig:overall_diagram}
    \vspace{-7mm}
\end{figure}


In this paper, we propose SHeRLoc, \underline{S}ynchronized \underline{He}terogeneous \underline{R}adar Place Recognition for Cross-Modal \underline{Loc}alization.
Despite the inherent challenges of heterogeneous radars, as illustrated in \figref{fig:overall_diagram}, SHeRLoc enables radar-agnostic learning through \ac{RCS} synchronization, where \ac{RCS} reflects the object's material and geometry.
By employing multi-view polar \ac{BEV}, SHeRLoc effectively addresses \ac{FOV} differences and achieves rotation invariance.
Additionally, we present HOLMES, a feature aggregation network that addresses radar speckle noise and multipath issues through a learnable weight matrix and ghostbin, while ensuring stable optimal transport with an adaptive entropy-regularized Sinkhorn algorithm.
Unlike high-dimensional single-scale descriptors \cite{arandjelovic2016netvlad, yao2024monocular}, which may lack global scene context, HOLMES integrates both local RCS patterns and global structural information into a compact descriptor.
Our FFT-similarity-based data mining and adaptive margin-based triplet loss enable spatially and viewpoint-aware metric learning.
By overcoming sparsity and noise patterns, SHeRLoc addresses the kidnapped robot problem with 4D radar on a spinning radar map, while also being applicable to \ac{FMCW} LiDAR.

Our contributions can be summarized as follows:
\begin{itemize}
    \item We present SHeRLoc, a deep network designed to tackle the complexities of heterogeneous radars, including sparsity, diverse \ac{FOV}, and varying data dimensions. To the best of our knowledge, this is the first approach specifically designed for heterogeneous radar systems.
    \item To address the modality gap, we propose a novel \ac{RCS} matching and multi-view polar projection, enabling radar-agnostic learning. Our preprocessing pipeline effectively removes dynamic objects, clutter, and ground return.
    \item We propose a hierarchical optimal transport-based feature aggregation method for rotationally robust descriptors, overcoming the loss of global context in traditional high-dimensional descriptors. Our FFT-similarity-based data mining and adaptive margin-based triplet loss ensure spatially and viewpoint-aware metric learning.
    \item We evaluate SHeRLoc against \ac{SOTA} methods under diverse scenarios (homogeneous, heterogeneous, single-/multi-session, and lidar-to-radar) and demonstrate superior performance in extreme conditions (e.g., crowded, snow, random rotation, and zero-shot). We open-source the code for the radar robotics community.

\end{itemize}

\section{related work}
\label{sec:relatedwork}
\subsection{Radar Place Recognition}

Spinning radar and \ac{SoC} radar are the two primary sensor types in radar PR.
Early radar PR mainly utilized spinning radar, leveraging their $360^\circ$ \ac{FOV} and relatively dense representations.
Radar Scan Context~\cite{kim2020mulran} extended the LiDAR-based Scan Context~\cite{kim2018scan} for rotation invariance, while Kidnapped Radar~\cite{suaftescu2020kidnapped} introduced a NetVLAD-based approach. 
As time efficiency became critical, 
RaPlace~\cite{jang2023raplace} and Open-RadVLAD~\cite{gadd2024open} adopted the Fourier Transform, and ReFeree~\cite{kim2024referee} proposed a compact one-dimensional descriptor.
These advancements established a foundation for radar PR but faced challenges when the sensor output is sparse, as seen in \ac{SoC} radar.

The adoption of \ac{SoC} radars in PR tasks has rendered traditional spinning radar-based methods less applicable due to the narrower \ac{FOV} and sparser data of \ac{SoC} radars, necessitating new approaches.
Notably, AutoPlace \cite{cai2022autoplace} increased the \ac{FOV} by using five automotive radars and overcame sparsity by utilizing a spatial-temporal encoder.
However, it relies on high-dimensional descriptors and requires an additional reranking process.
mmPlace \cite{meng2024mmplace} expands the \ac{FOV} by leveraging a rotating platform and concatenating heatmaps, while SPR \cite{herraez2024spr} proposes a lightweight method using a single radar scan. However, both methods are vulnerable in dynamic environments as they ignore velocity information.
Recently, TransLoc4D~\cite{peng2024transloc4d} employs transformer architectures to integrate geometry, intensity, and velocity from 4D radar, but remains susceptible to rotation due to its reliance on a single viewpoint.
\citet{hilger2025introspective} considered both co- and counter-directional viewpoints; nevertheless, this approach is still limited to just two viewpoints, and the challenge of PR across heterogeneous radar with distinct data characteristics remains unsolved.
\subsection{Challenges in Heterogeneous Sensor Place Recognition}

Heterogeneous sensor PR requires seamless integration of sensors with distinct data characteristics.
RaLF~\cite{nayak2024ralf} retrieves spinning radar queries from existing LiDAR maps, while Get to the Point~\cite{tang2021get} leverages overhead imagery for LiDAR PR and metric localization. However, both approaches struggle to handle differing \ac{FOV}s.
Similarly, \citet{cattaneo2020global} proposed a LiDAR-Camera PR model without \ac{FOV} considerations, resulting in $360^\circ$ point clouds that inevitably include irrelevant scene information outside the RGB camera’s \ac{FOV}.
ModaLink \cite{xie2024modalink} addressed this by using camera pose priors to crop point clouds, but this cropping approach is impractical since the required extrinsic calibration is rarely available.
In response, LCPR \cite{zhou2023lcpr} integrated sensor data using panoramic views, while \citet{yao2024monocular} introduced xNetVLAD that accounts for \ac{FOV} differences. However, these methods require separate networks for each modality.
Recently, HeLiOS \cite{jung2025helios} introduced overlap-based learning with guided-triplet loss for heterogeneous LiDAR PR, though it is limited by LiDAR's high point density and minimal differences between heterogeneous LiDAR scans. 
In contrast to previous methods, we propose a novel RCS polar matching to achieve sensor-agnostic learning, overcoming radar's sparsity and distinct noise patterns.
\begin{figure*}[!t]
    \centering
    \includegraphics[trim= 2cm 1.8cm 2cm 1.65cm, clip,width=1.0\textwidth]{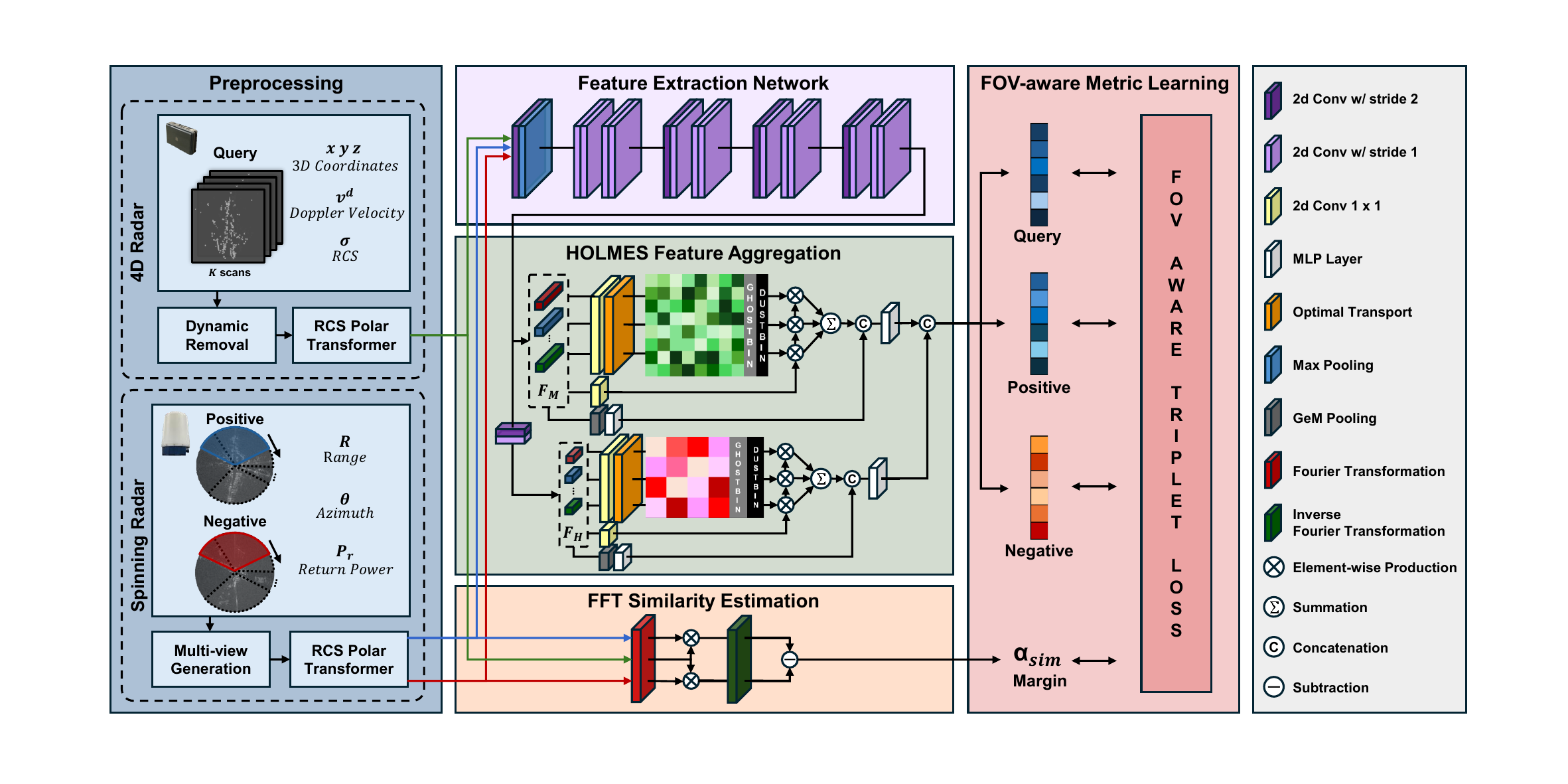}
    \caption{The overall pipeline of SHeRLoc. \ac{RCS} polar images \(I_{\text{4D}}\) and \(I_{\text{spin}}\) are generated from heterogeneous radars and processed through a shared feature extraction network \(\mathcal{G}\). Multi-level features \(F_M\) and \(F_H\) are aggregated into global descriptors \(\mathcal{D}\) using HOLME.}
    \label{fig:pipeline}
    \vspace{-6mm}
\end{figure*}

\section{Methods}
\label{sec:methods}


As illustrated in \figref{fig:pipeline}, SHeRLoc transforms 4D radar scans \(\mathcal{S}_\text{4D}\) and spinning radar scans \(\mathcal{S}_{\text{spin}}\) into \ac{RCS} polar BEV images denoted as \(I_\text{4D}, I_{\text{spin}} \in \mathbb{R}^{H \times W}\). Here, \(H\) corresponds to a maximum range \(\rho_{\text{max}}\), and \(W\) corresponds to an azimuth span \(\phi\). The synchronized data is then processed by a feature extraction network \(\mathcal{G}\), and local features \(F_\text{4D}, F_{\text{spin}}\) are aggregated into global descriptors \(\mathcal{D}_\text{4D}, \mathcal{D}_\text{spin}\) using HOLMES \(\mathcal{H}\), respectively.


\subsection{Preprocessing}
\subsubsection{Dynamic Removal with Ego-velocity Estimation}
Each 4D radar scan at time \(k\), \(\mathcal{S}_{\text{4D},k} = \{(x_i, y_i, z_i, v_i^d, \sigma_i)\}_{i=1}^N \in \mathbb{R}^{N \times 5}\), comprises \(N\) points with 3D coordinates \((x_i, y_i, z_i)\), Doppler velocity \(v_i^d\), and \ac{RCS} \(\sigma_i\).
We employ the 3-Point RANSAC-LSQ~\cite{doer2020ekf} to estimate the ego-velocity and apply a refinement function \(f_{\text{removal}}\), which filters out moving objects, clutter, and ground returns. 
The refined scan is defined as
\(\mathcal{S}'_{\text{4D},k} = f_{\text{removal}}(\mathcal{S}_{\text{4D},k}) 
= \{ (x_i, y_i, z_i, v_i^d, \sigma_i) \in \mathcal{S}_{\text{4D},k} \mid v_i^d < \tau_v,\; z_i \geq \tau_z,\; \sigma_i \geq \tau_\sigma \},\)
where each \(\tau\) denotes a threshold.

\subsubsection{\ac{FOV} Matching with Multi-view Polar Projection}
Prior works \cite{shi2023panovpr, yao2024monocular} address varying \ac{FOV} issues in panoramic front-view settings using sliding window cropping. However, such approaches are unsuitable for spinning radar lacking elevation measurements.
Additionally, the 4D radar has a sector-shaped \ac{FOV} in Cartesian \ac{BEV}, leaving empty regions in the rectangular image and thereby degrading performance. 
In contrast, polar \ac{BEV} provides a consistent rectangular representation of overlapping regions between heterogeneous radars.

For the 4D radar, \(\mathcal{S}'_{\text{4D},k}\) is projected into \ac{RCS} polar \ac{BEV} image \(I_{\text{4D},k} \in \mathbb{R}^{H \times W}\) using a polar mapping function \(f_{\text{polar}}: \mathcal{S}'_{\text{4D},k} \rightarrow I_{\text{4D},k}(r, \theta)\), where \(r\) and \(\theta\) are defined as:
\vspace{-1mm}
\begin{equation}\small
r =\left[\sqrt{x^2 + y^2}\right] \frac{H}{\rho_{max}}, \quad \theta = \left[ 1 - 2 \phi^{-1} \arctan(\frac{y}{x}) \right]\frac{W}{2}.
\end{equation}
\normalsize
Each pixel in \(I_{\text{4D},k}\) encodes \ac{RCS} in a scaled form, consistent with the spinning radar convention of quantizing return power in half-dB steps (e.g., 100 corresponds to 50 dB).
To form a temporally aggregated 4D radar image \(I_{\text{4D},t}\) at time \(t\), we apply max-pooling over \(K\) consecutive frames:
\vspace{-1mm}
\begin{equation}\small
I_{\text{4D},t}(r, \theta) = \max_{k \in \{t - K + 1, \dots, t\}} I_{\text{4D},k}(r, \theta).
\label{eqn:maxpool}
\end{equation}
\normalsize

For the spinning radar, we rescale \(\mathcal{S}_{\text{spin},t} \in \mathbb{R}^{N_{\text{r}} \times N_{\text{a}}}\), with \(N_r\) range bins and \(N_a\) azimuth bins, to \(\mathcal{S}'_{\text{spin},t} \in \mathbb{R}^{H \times 3W}\) to account for its three times larger \ac{FOV}.
We then extract \(n_v\) sub-views \(\{M_{\text{spin},t}^j\}_{j=1}^{n_v}\) from \(\mathcal{S}'_{\text{spin},t}\) using a sliding window of size \(H \times W\) along the azimuthal axis. Each sub-view with azimuth offset \(\Delta\)  between consecutive views is given by:
\vspace{-1mm}
\begin{equation}\small
    M_{\text{spin},t}^j = \mathcal{S}'_{\text{spin},t}[:,\, \Delta (j-1) : \Delta (j-1) + W].
\end{equation}
\normalsize
This multi-view generation function \(f_{\text{multiview}}: \mathcal{S}'_{\text{spin},t} \rightarrow \{M_{\text{spin},t}^j \in \mathbb{R}^{H \times W}\}_{j=1}^{n_v}\) ensures $360^\circ$ \ac{FOV} coverage, supporting rotation-invariance, detailed in Section~\ref{sec: Rotation Invariant}.

\subsubsection{\ac{RCS} Matching}
We match \ac{RCS} values \(\sigma\) with return power \(P_r\) using the classical radar equation, formulated as:
\vspace{-1mm}
\begin{equation}\small
    P_r = \frac{P_t \cdot G^2 \cdot \lambda^2 \cdot \sigma}{(4\pi)^3 \cdot R^4},
    \label{eqn:radar_equation}
\end{equation}
\normalsize
where \(P_t\) is the transmitted power, \(G\) is the antenna gain, \(\lambda\) is the wavelength, and \(R\) is the range.
An object at height \(h\) above the ground and slant range \(R\) forms an angle \(\alpha\) given by \(\sin \alpha = h / R\), or equivalently, \(R = h \csc \alpha\). Inspired by air-surveillance radars that maintain constant received power regardless of range~\cite{hao2017developing}, modern spinning radars adopt a cosecant-squared beam profile, where the antenna gain satisfies \( G\propto R^2\). Assuming constant \(P_t\) and \(\lambda\), the \ac{RCS} in the radar equation~\eqref{eqn:radar_equation} is simplified on the dBsm scale as:
\vspace{-1mm}
\begin{equation}\small
\sigma_{\text{dBsm}} = P_r \, [\text{dB}] + C, \label{eqn:rcs}
\end{equation}
\normalsize
where \(\sigma_{\text{dBsm}} = 10 \log_{10} (\sigma)\), with the reference \ac{RCS} defined as \(\sigma_{\text{ref}} = 1 \, \text{m}^2\), and \(C = 40 \log_{10} (R) + 30 \log_{10} (4 \pi) - 20 \log_{10} (G \cdot \lambda) - P_t \, [\text{dB}] \) is a correction constant.

To synchronize \(M_{\text{spin}, t}\) with \(I_{\text{4D}, t}\), where pixels represent \(P_r\) and \(\sigma\) in half-dB steps, respectively, we precompute a correction term \(C_{corr}\) only once per radar pair to align \(I_{\text{4D}, i}\) with the most similar subview \(M_{\text{spin}, i}^{\text{max}}\).
Then, each \(M_{\text{spin}, t}^j\) is processed by an \ac{RCS} matching module \(f_{\text{rcs}}: M_{\text{spin},t}^j \rightarrow I_{\text{spin},t}^j\) to produce the final \ac{RCS} polar \ac{BEV} representation:
\vspace{-1mm}
\begin{equation}\small
    I_{\text{spin},t}^j = f_{\text{rcs}}(M_{\text{spin},t}^j) = M_{\text{spin},t}^j + C_{corr},
\end{equation}
\normalsize
where \(C_{corr}\) is obtained by optimizing a per-frame correction term \(k_i\) to minimize the Huber loss. Let the set of valid pixel coordinates be:
\(\Omega_i = \{ (r,\theta) \mid M_{\text{spin,i}}^{\text{max}}(r,\theta) \neq 0 \text{ and } I_{\text{4D},i}(r,\theta) \neq 0 \}\),
and define the Huber loss as:
\vspace{-1mm}
\begin{equation}\small
    L_\delta(r) =
    \begin{cases}
        \frac{1}{2}r^2, & \text{if } |r| \le \delta, \\
        \delta \left( |r| - \frac{1}{2}\delta \right), & \text{if } |r| > \delta.
    \end{cases}
\end{equation}
\normalsize
Then, The loss for each image pair \(i\) can be formulated as:
\vspace{-1mm}
\begin{equation}\small
L_i(k_i) = \frac{1}{|\Omega_i|} \sum_{(r, \theta) \in \Omega_i} L_\delta \bigl( I_{\text{4D},i}(r, \theta) - (M_{\text{spin},i}^\text{max}(r, \theta) + k_i) \bigr).
\end{equation}
\normalsize
To ensure smooth variation of \(k_i\), we introduce a regularization term with factor \(\lambda\) in the joint optimization problem:
\vspace{-1mm}
\begin{equation}\small
\label{eqn:k}
\{ k_i^* \}_{i=1}^N = 
\mathop{\mathrm{argmin}}_{\displaystyle \{ k_i \}_{i=1}^N} 
\left( \sum_{i=1}^N L_i(k_i) + \lambda \sum_{j=2}^N (k_j - k_{j-1})^2 \right).
\end{equation}
\normalsize

\begin{figure}[!t]
    \centering    \includegraphics[trim= 3.1cm 2.9cm 3.1cm 2.7cm, clip,width=1.0\columnwidth]{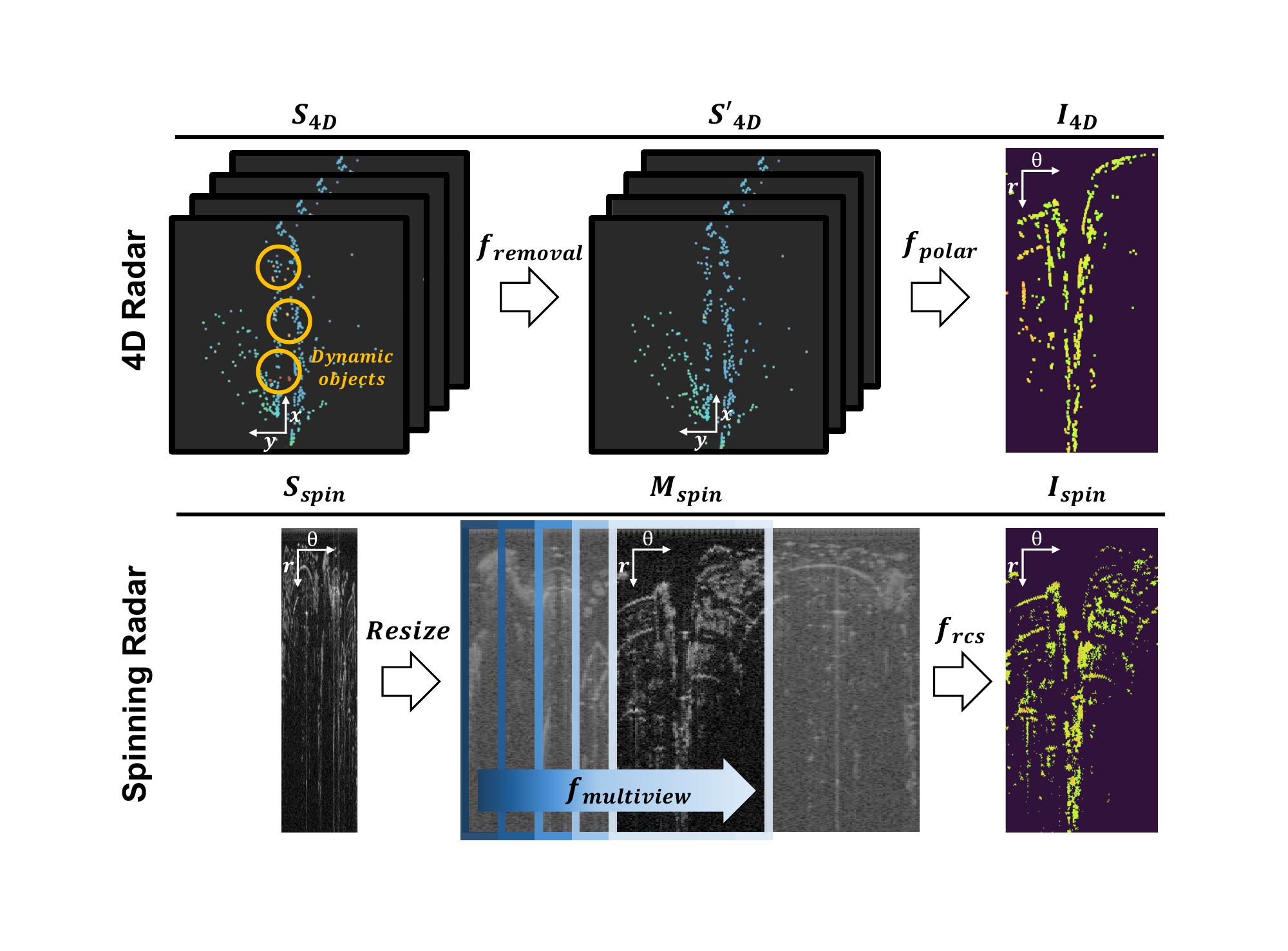}
    \caption{The pipeline of RCS polar synchronization. The turbo colormap is applied to \(I_{\text{4D}}\) and \(I_{spin}\) for visualization clarity.}
    \label{fig:rcs}
    \vspace{-4mm}
\end{figure}
Then, we compute \(C_{corr} = \frac{1}{N} \sum_{i=1}^N k_i^*\). 
As shown in \figref{fig:rcs}, 4D radar scans \(\mathcal{S}_\text{4D}\) and spinning radar scans \(\mathcal{S}_{\text{spin}}\) are synchronized into an RCS polar \ac{BEV} representation.

\subsection{Feature Extraction Network}
The preprocessed $I_{\text{4D}}, I_{\text{spin}} \in \mathbb{R}^{H \times W}$ are fed simultaneously into a ResNet-based architecture~\cite{he2016deep}, \(\mathcal{G} : I_{\text{4D}}, I_{\text{spin}} \rightarrow F_{\text{4D}}, F_{\text{spin}}\), which generate the feature maps $F_{\text{4D}}, F_{\text{spin}} \in \mathbb{R}^{C \times H/32 \times W/32}$.
Traditional heterogeneous PR methods require separate encoders for each sensor \cite{nayak2024ralf} or a backbone network for independent learning  \cite{cattaneo2020global, yao2024monocular}.
However, due to the shared RCS polar BEV format, our approach allows the use of a common network across radars, unlike previous methods. 

\subsection{Multi-scale Descriptors with Optimal Transport}

To achieve a compact integration of complementary feature hierarchies, we propose HOLMES, \underline{H}ierarchical \underline{O}ptimal transport for \underline{L}ocally aggregated \underline{M}ulti-scale descriptors with adaptive \underline{E}ntropy-regularized \underline{S}inkhorn-algorithm.
Previously, SALAD~\cite{izquierdo2024optimal} fuses local features with a DINOv2 CLS token, while HeLiOS~\cite{jung2025helios} combines them with GeM-pooled global features. However, both result in high-dimensional descriptors, whereas HeLiOS-S, a smaller variant of HeLiOS, sacrifices expressiveness and global context.
We address these issues with a hierarchical framework that combines local RCS patterns and global context. Mid-level features form a local feature matrix $\mathbf{F_M} \in \mathbb{R}^{n_M \times d_{f_M}}$, and high-level features, processed through an additional CNN layer, form $\mathbf{F}_H\in \mathbb{R}^{{n_H} \times {d_{f_H}}}$, where $n$ is the number of spatial locations and $d_f$ is the feature dimension. 
A convolution layer predicts a score matrix $\mathbf{S} \in \mathbb{R}^{n_M \times m_M}$ for \(\mathbf{F_M}\), where $m_M$ denotes the number of clusters.
Inspired by GhostVLAD~\cite{zhong2019ghostvlad}, we extend SALAD’s dustbin with a \emph{ghostbin} to handle noise points, resulting in a score matrix $\bar{\mathbf{S}} \in \mathbb{R}^{n_M \times (m_M+2)}$.
The Sinkhorn algorithm~\cite{cuturi2013sinkhorn} optimizes feature-to-cluster assignments, yielding a refined matrix $\mathbf{R} \in \mathbb{R}^{n_M \times m_M}$ by normalizing $\exp(\bar{\mathbf{S}})$ and dropping dustbin and ghostbin.
We introduce an adaptive entropy regularization as:

\small
\vspace{-1.5mm}
\small
\begin{equation}
\text{reg} = 1 + 2 \cdot \tanh \left( \frac{\frac{1}{N} \sum_{i=1}^{N} (x_i - \mu)^2}{2(\mu + \epsilon)} \right),
\end{equation}
\normalsize
where \( x_i \) is \ac{RCS} feature, \( \mu \) is the mean and \( N \) is the number of elements. We compute the variance relative to \(\mu\) and apply \(tanh(\cdot)\) to bound the regularization smoothly. High variance induces smooth matching for stability, while low variance leads to precise matching. Then, the aggregated feature matrix $\mathbf{V} \in \mathbb{R}^{m_M \times l_M}$ with the cluster dimension $l_M$ is computed as:

\vspace{-2.5mm}
\small
\begin{equation}
V_{j,k} = \sum_{i=1}^n R_{i,k} \cdot \bar{F}_{i,j},
\end{equation}
\normalsize
where $\mathbf{\bar{F}}$ is the feature derived by applying a dimensionality reduction to $\mathbf{F}$.
Meanwhile, GeM pooling and MLP layers produce a global representation $\mathbf{G} \in \mathbb{R}^{s_M}$. Concatenating $\mathbf{V}$ and $\mathbf{G}$ yields $\mathbf{g} \in \mathbb{R}^{m_M \cdot l_M + s_M}$, which is then transformed by a learnable weight matrix $\mathbf{H} \in \mathbb{R}^{(m_M \cdot l_M + s_M) \times d_M}$ into the mid-level descriptor $\mathcal{D}_M \in \mathbb{R}^{d_M}$. Applying the same process to \(\mathbf{F}_H\) produces a high-level descriptor \(\mathcal{D}_H \in \mathbb{R}^{d_H}\). 
By concatenating the mid-level descriptor, the HOLMES \(\mathcal{H}\) generates the final descriptor \(\mathcal{D}=\mathcal{D}_M \oplus \mathcal{D}_H \in \mathbb{R}^{d_M + d_H}\).

\subsection{\ac{FOV}-aware Metric Learning}
\subsubsection{\ac{FOV}-aware Data Mining}
We introduce a \ac{FOV}-aware data mining approach that selects positive and negative samples based on view similarity. For efficient similarity computation, we leverage the Convolution Theorem, performing computations in the frequency domain \cite{duhamel1990fast}. Concretely, the cross-correlation between two scans \(I_1\) and \(I_2\) is defined as:
\vspace{-1.5mm}
\begin{equation}\small
\text{Sim}(I_1, I_2) = 
\frac{
\max \left[ 
\mathcal{F}^{-1} \left( 
\mathcal{F}(I_1) \cdot \overline{\mathcal{F}(I_2)} 
\right)
\right]
}{
\|I_1\|_2 \cdot \|I_2\|_2
},
\end{equation}
\normalsize
where \(\mathcal{F}\) and \(\mathcal{F}^{-1}\) denote the Fourier and inverse Fourier transforms, respectively. The complexity is reduced from \(\mathcal{O}(N^2)\) to \(\mathcal{O}(N \log N)\), making data mining time-efficient.

We select the multi-view set \(\{I_{\text{spin},t}^j\}_{j=1}^{n_v}\) with the closest timestamp to the corresponding \(I_{\text{4D},t}\). Among these multi-views, the scan \(I_{\text{spin},t}^{\text{max}}\) that has the highest similarity is labeled as a positive. In contrast, scans beyond the 25\,m radius are considered negative, even if their \ac{FOV} partially overlap, ensuring both spatially and viewpoint-wise meaningful.

\subsubsection{\ac{FOV}-aware Triplet Loss}
To train the network with the mined samples, we adopt a FoV-aware triplet loss formulation. The objective is to bring positive samples closer in the embedding space while pushing negatives apart, with a margin that dynamically adapts to the degree of similarity:

\vspace{-5.0mm}
\begin{equation}\small
\label{eqn:triplet}
\mathcal{L}_{\text{triplet}} = \max \left( d(x_{i}^{q}, x_{j}^{p}) - \min_{n=1}^{N} d(x_{i}^{q}, x_{j}^{n}) + \alpha_{\text{sim}}, 0 \right),
\end{equation}
\normalsize
where \(x_{i}^{q}\) is the query pair, \(x_{j}^{p}\) is the positive pair, \(x_{j}^{n}\) are negative pair, and \(d(\cdot)\) denotes the  \(L_2\) distance in the embedding space. The adaptive margin \(\alpha_{\text{sim}}\) is defined as:

\vspace{-3mm}
\begin{equation}\small
\label{eqn:alpha}
\alpha_{\text{sim}} = \gamma \left( \text{Sim}(\mathbf{q}, \mathbf{p}) - \text{Sim}(\mathbf{q}, \mathbf{n}) \right),
\end{equation}
\normalsize
where \(\gamma\) is a scale factor. By incorporating similarity into the margin, the loss becomes sensitive to \ac{FOV} overlap. If a negative has a similar view to the query, the margin becomes smaller, making the model focus on harder negatives.

\subsection{Rotation Invariance and Intentional Translation Variance}
\label{sec: Rotation Invariant}
\begin{figure}[!t]
    \centering    \includegraphics[trim= 5.7cm 3.8cm 5.4cm 4.5cm, clip, width=1.0\columnwidth]{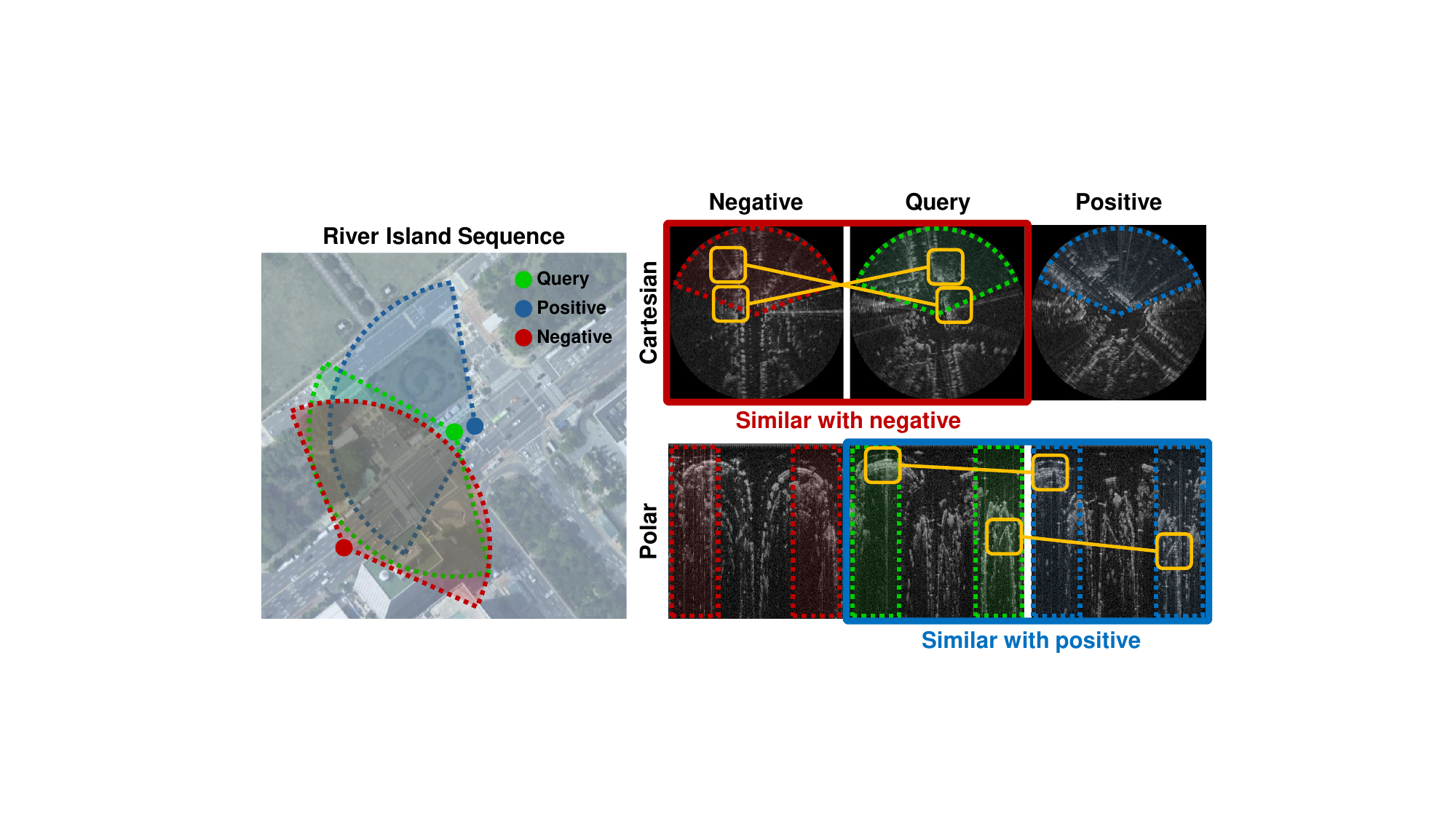}
    \caption{For a far but overlapping negative and a nearby but rotated positive, Cartesian BEV yields higher similarity with the negative due to rotation variance. In contrast, polar BEV produces higher similarity with the positive, demonstrating robustness to rotation.}
    \label{fig: qpn}
    \vspace{-6mm}
\end{figure}

As shown in \figref{fig: qpn}, spatially distant negatives with overlapping \ac{FOV} produce similar features when using Cartesian \ac{BEV}, as seen in \cite{nayak2024ralf, cai2022autoplace}. Also, nearby positives may exhibit feature mismatches due to small rotations. 
To address this, we leverage \(f_{\text{polar}}\),  \(f_{\text{multiview}}\), \(\mathcal{G}\), and \(\mathcal{H}\) to achieve rotation invariance and intentional translation variance, thereby maintaining discriminative power in long-range radar PR.
See RING++~\cite{xu2023ring++} for more details about invariance and equivariance.

\subsubsection{Pipeline Invariance}
A rotated scan \(I'(r, \theta) = I(r, \theta + \delta \theta)\) shifts multi-view azimuths cyclically.
Then, \(\mathcal{G}\) outputs \(F'_j(r, \theta)\), and \(\mathcal{H}\) produces an invariant descriptor.
With polar \ac{BEV} ensuring translation variance, the pipeline is rotation-invariant and intentionally translation variant, satisfying \(\mathcal{H}(\mathcal{G}(f_{\text{multiview}}(T_{\delta \theta} S(r, \theta)) = \mathcal{H}(\mathcal{G}(f_{\text{multiview}}(S(r, \theta)))\).

\subsubsection{Feature Extraction Equivariance}

Since a rotation transformation \(T_{\delta \theta}\) corresponds to a shift in the azimuth dimension and \(f_{\text{multiview}}\) operates uniformly across \(\theta\), the output of \(f_{\text{multiview}}\) is shifted by the same \(\delta \theta\):
\(f_{\text{multiview}}(T_{\delta \theta} S(r, \theta))= T_{\delta \theta} f_{\text{multiview}}(S(r, \theta)).\)
Similarly, the CNN-based network \(\mathcal{G}\) preserves translation-equivariance up to edge effects \cite{cohen2016group}. Therefore, both \(f_{\text{multiview}}\) and \(\mathcal{G}\) maintain the translation-equivariance property. Theoretically, we have to generate infinite multi-views, and aliasing may occur due to downsampling. However, robust performance is achieved with a limited number of multi-views, detailed in Section~\ref{sec: Robustness}.

\subsubsection{HOLMES Invariance}
\(\mathcal{H}\) processes feature maps \(F\) to produce \(\mathcal{D}=\mathcal{D}_M \oplus \mathcal{D}_H\).
GeM Pooling produces a global feature \(\mathbf{G}\), invariant to translations, as the summation is order-independent.
A convolutional layer produces a score matrix \(\mathbf{S}\) for Optimal Transport, and the Sinkhorn algorithm computes assignments \(\mathbf{R}\). 
The aggregated feature \(\mathbf{V}_{j,k} = \sum_{i=1}^n R_{i,k} \cdot F_{i,j}\) is invariant, as summation reorders with translated inputs \(F'_j(r, \theta)\).
Ghostbin and dustbin discard non-informative points, preserving invariance by excluding spatial dependencies.
For multi-scale processing, high-level features follow the same process.
Concatenation (\(\mathcal{D}_M \oplus \mathcal{D}_H\)) remains invariant to cyclic shifts. For translated feature maps \(F'_j(r, \theta)\), \(\mathcal{H}\) yields \( \mathbf{D}'_M \oplus \mathbf{D}'_H = \mathcal{H}(F'_j(r, \theta)) = \mathcal{H}(F_j(r, \theta)) = \mathcal{D}_M \oplus \mathcal{D}_H.\)
Thus, our proposed HOLMES network is translation-invariant.

\definecolor{mycolor1}{HTML}{def3e6}
\definecolor{mycolor2}{HTML}{ecf8f1}

\begin{table*}[t]
\caption{Performance Comparison for Single-session Place Recognition with Heterogeneous Radar in Various Challenging Conditions \\ (\faSun: Clear, \faCloudSun: Dusk, $\rightmoon$: Night, \faCloud: Cloud, \faSnowflake[regular]: Snow, \faCarSide: dynamic object-rich, \colorbox{mycolor1}{\textbf{Bold}}: Best, \colorbox{mycolor2}{\underline{Underline}}: Second Best)}
\label{tab: Hetero_single}
\centering
\resizebox{\textwidth}{!}{
\begin{tabular}{l|cccccccccccccccccc} 
\toprule

 &
  \multicolumn{6}{c}{\texttt{Sports Complex}} &
  \multicolumn{6}{c}{\texttt{Library}} &
  \multicolumn{6}{c}{\texttt{River Island}} \\
 &
  \multicolumn{2}{c}{01 \faSun} &
  \multicolumn{2}{c}{02 $\rightmoon$} &
  \multicolumn{2}{c}{03 \faSnowflake[regular]} &
  \multicolumn{2}{c}{01 \faSun} &
  \multicolumn{2}{c}{02 $\rightmoon$} &
  \multicolumn{2}{c}{03 \faSnowflake[regular]} &
  \multicolumn{2}{c}{01 \faSun\ \faCarSide } &
  \multicolumn{2}{c}{02 \faCloudSun\ \faCloud\ \faCarSide } &
  \multicolumn{2}{c}{03 \faCloud\ \faCarSide} 
  \\ \cline{2-19} \rule{0pt}{2.5ex}
  
  \multirow{-3}{*}{Methods} & 
  \multicolumn{1}{l}{R@1} &
  \multicolumn{1}{l}{R@1\%} &
  \multicolumn{1}{l}{R@1} &
  \multicolumn{1}{l}{R@1\%} &
  \multicolumn{1}{l}{R@1} &
  \multicolumn{1}{l}{R@1\%} &
  \multicolumn{1}{l}{R@1} &
  \multicolumn{1}{l}{R@1\%} &
  \multicolumn{1}{l}{R@1} &
  \multicolumn{1}{l}{R@1\%} &
  \multicolumn{1}{l}{R@1} &
  \multicolumn{1}{l}{R@1\%} &
  \multicolumn{1}{l}{R@1} &
  \multicolumn{1}{l}{R@1\%} &
  \multicolumn{1}{l}{R@1} &
  \multicolumn{1}{l}{R@1\%} &
  \multicolumn{1}{l}{R@1} &
  \multicolumn{1}{l}{R@1\%} \\ \midrule

Radar SC~\cite{kim2020mulran}
& 0.047
& 0.135
& 0.042
& 0.061
& 0.044
& 0.064
& 0.040
& 0.060
& 0.041
& 0.063
& 0.042
& 0.066
& 0.010
& 0.020
& 0.004
& 0.012
& 0.005
& 0.030 \\

RaPlace~\cite{jang2023raplace}
& 0.042
& 0.148
& 0.056
& 0.156
& 0.025
& 0.152
& 0.035
& 0.069
& 0.040
& 0.090
& 0.016
& 0.123
& 0.002
& 0.152
& 0.000
& 0.024
& 0.004
& 0.162\\

RadVLAD~\cite{gadd2024open}
& 0.019
& 0.334
& 0.051
& 0.222
& 0.022
& 0.426
& 0.011
& 0.237
& 0.014
& 0.295
& 0.012
& 0.245
& 0.001
& 0.217
& 0.066
& 0.543
& 0.006
& 0.367\\ 
FFT-RadVLAD~\cite{gadd2024open}
& 0.040
& 0.344
& 0.014
& 0.186
& 0.022
& 0.377
& 0.023
& 0.287
& 0.014
& 0.273
& 0.011
& 0.229
& 0.002
& 0.330
& 0.026
& 0.533
& 0.001
& 0.488\\

Autoplace~\cite{cai2022autoplace}
& 0.015
& 0.198
& 0.046
& 0.098
& 0.017
& 0.186
& 0.010
& 0.129
& 0.031
& 0.131
& 0.018
& 0.122
& 0.003
& 0.051
& 0.001
& 0.009
& 0.001
& 0.021\\

\midrule
  \cellcolor[HTML]{f3f7fc}SHeRLoc-S
& \cellcolor[HTML]{ecf8f1}\underline{0.857}
& \cellcolor[HTML]{ecf8f1}\underline{0.924}
& \cellcolor[HTML]{def3e6}\textbf{0.975}
& \cellcolor[HTML]{ecf8f1}\underline{0.985}
& \cellcolor[HTML]{ecf8f1}\underline{0.945}
& \cellcolor[HTML]{def3e6}\textbf{0.981}
& \cellcolor[HTML]{ecf8f1}\underline{0.866}
& \cellcolor[HTML]{ecf8f1}\underline{0.917}
& \cellcolor[HTML]{ecf8f1}\underline{0.925}
& \cellcolor[HTML]{ecf8f1}\underline{0.950}
& \cellcolor[HTML]{ecf8f1}\underline{0.850}
& \cellcolor[HTML]{ecf8f1}\underline{0.891}
& \cellcolor[HTML]{def3e6}\textbf{0.899}
& \cellcolor[HTML]{def3e6}\textbf{0.965}
& \cellcolor[HTML]{def3e6}\textbf{0.868}
& \cellcolor[HTML]{def3e6}\textbf{0.963}
& \cellcolor[HTML]{ecf8f1}\underline{0.850}
& \cellcolor[HTML]{ecf8f1}\underline{0.945}
\\

  \cellcolor[HTML]{f3f7fc}SHeRLoc
& \cellcolor[HTML]{def3e6}\textbf{0.900}
& \cellcolor[HTML]{def3e6}\textbf{0.936}
& \cellcolor[HTML]{ecf8f1}\underline{0.962}
& \cellcolor[HTML]{def3e6}\textbf{0.987}
& \cellcolor[HTML]{def3e6}\textbf{0.958}
& \cellcolor[HTML]{ecf8f1}\underline{0.980}
& \cellcolor[HTML]{def3e6}\textbf{0.881}
& \cellcolor[HTML]{def3e6}\textbf{0.936}
& \cellcolor[HTML]{def3e6}\textbf{0.938}
& \cellcolor[HTML]{def3e6}\textbf{0.964}
& \cellcolor[HTML]{def3e6}\textbf{0.868}
& \cellcolor[HTML]{def3e6}\textbf{0.912}
& \cellcolor[HTML]{ecf8f1}\underline{0.880}
& \cellcolor[HTML]{ecf8f1}\underline{0.957}
& \cellcolor[HTML]{ecf8f1}\underline{0.860}
& \cellcolor[HTML]{ecf8f1}\underline{0.952}
& \cellcolor[HTML]{def3e6}\textbf{0.858}
& \cellcolor[HTML]{def3e6}\textbf{0.959}
\\

\bottomrule
\end{tabular}
}
\vspace{-4mm}
\end{table*}



\section{Experimental Results}
\label{sec:experiment}

\subsection{Implementation Details}
We trained SHeRLoc on a GeForce RTX 3090 using the AdamW optimizer with CosineAnnealingLR scheduler, starting with a learning rate of 1e-3 and decaying to a minimum of 1e-5 over 20 epochs. 
For the polar BEV projection, the resolution of a polar image with a maximum range $\rho_{\text{max}}=150\,\text{m}$ and an azimuth span $\phi=120^\circ$ is $384 \times 192$.
The number of scans K is set to 5 for \eqref{eqn:maxpool}; the multi-view offset \(\Delta\) is set to $10^\circ$(16 pixels); and \(\lambda\) is set to 0.1 for \eqref{eqn:k}. For HOLMES, \((m_M, l_M, s_M, d_M)\) are set to \((64,\ 256,\ 256,\ 256)\), and \((m_H, l_H, s_H, d_H)\) are set to \((16,\ 64,\ 64,\ 64)\), respectively.
We utilize one positive and five negative samples for \eqref{eqn:triplet}, and \(\gamma\) is set to 1.0 for \eqref{eqn:alpha}.
For SHeRLoc-S, a lightweight variant, only the mid-level descriptor \(\mathcal{D}_M \in \mathbb{R}^{256}\) is retained.




\subsection{Heterogeneous Radar Dataset and Evaluation Metrics}
To evaluate SHeRLoc, a dataset must include both 4D radar and spinning radar. Existing public datasets, such as Oxford Radar RobotCar~\cite{barnes2020oxford} and MulRan~\cite{kim2020mulran}, are limited to spinning radar, while NTU4DRadLM~\cite{zhang2023ntu4dradlm} and SNAIL Radar~\cite{huai2024snail} focus solely on 4D radar. 
In contrast, HeRCULES~\cite{kim2025hercules} is the only publicly available heterogeneous range sensor dataset, integrating 4D radar, spinning radar, and FMCW LiDAR, making it uniquely suited for heterogeneous radar PR evaluation.

We evaluate performance using Recall@K (R@K) and Average Recall@K (AR@K). A retrieval is deemed correct if the predicted location lies within a 5\,m radius of the ground-truth. In addition, we employ the \ac{PR} curve to assess overall retrieval performance. The PR curve is obtained by sweeping a threshold $\tau$ over the descriptor distance $d(x_i^q, x_j^m)$ between query and map embeddings. For each $\tau \in [0,2]$, uniformly sampled at 1000 intervals, a retrieved pair is considered a match if $d(x_i^q, x_j^m) < \tau$.

\subsection{Comparison with State-of-the-Art Methods}
As the first heterogeneous radar PR model, SHeRLoc was compared against homogeneous PR \ac{SOTA} models, including spinning radar PR methods Radar Scan Context~\cite{kim2020mulran}, RaPlace~\cite{jang2023raplace}, RadVLAD, and FFT-RadVLAD~\cite{gadd2024open}, as well as \ac{SoC} radar PR models Autoplace~\cite{cai2022autoplace}, TransLoc4D~\cite{peng2024transloc4d}, and LiDAR PR method MinkLoc3Dv2~\cite{komorowski2022improving}. For fair comparison, we used 5-scan aggregation for 4D radar and applied \ac{RCS} matching for spinning radar across all models.
Since methods like RaPlace~\cite{jang2023raplace} and AutoPlace~\cite{cai2022autoplace} operate on BEV images, they were included in the heterogeneous radar PR evaluation.
Also, all network-based models were trained on the \texttt{Mountain 01-03}, \texttt{Bridge 01}, \texttt{Stream 02}, and \texttt{Parking Lot 03-04}, while validation was conducted on \texttt{Parking Lot 01-02} of HeRCULES.

\subsubsection{Heterogeneous Radar Single-session PR}
\begin{figure}[t!]
    \centering
    \includegraphics[trim= 7.5cm 3cm 7.4cm 5cm, clip, width=1.0\columnwidth]{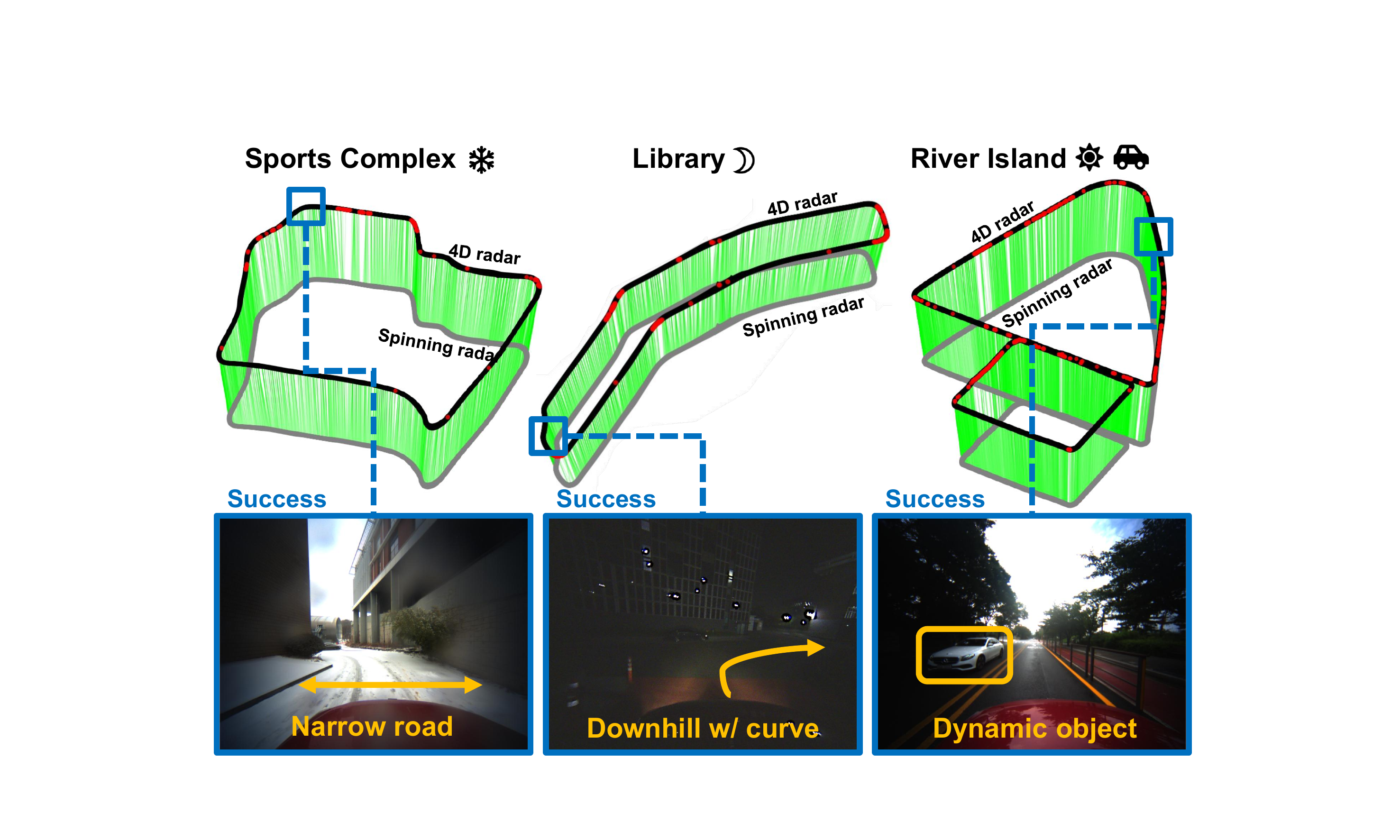}
    \caption{Trajectory from \texttt{Sports Complex}, \texttt{Library}, and \texttt{River Island} sequences, with green indicating true matching pairs, highlighting SHeRLoc's robustness in challenging scenarios.}
    \label{fig: match}
    \vspace{-4mm}
\end{figure}

For single-session evaluation, spinning radar data were used as the database and 4D radar data as queries across the \texttt{Sports Complex}, \texttt{Library}, and \texttt{River Island} sequences. As shown in \tabref{tab: Hetero_single}, SHeRLoc consistently outperforms all baselines, while the lightweight SHeRLoc-S also surpasses all \ac{SOTA} methods. 
Furthermore, as illustrated in \figref{fig: match}, SHeRLoc achieves robust performance even in challenging scenarios such as night, snow, and dynamic object-rich environments.

\begin{figure}[t!]
    \centering
    \includegraphics[trim= 9.2cm 5cm 9.3cm 5.3cm, clip, width=1.0\columnwidth]{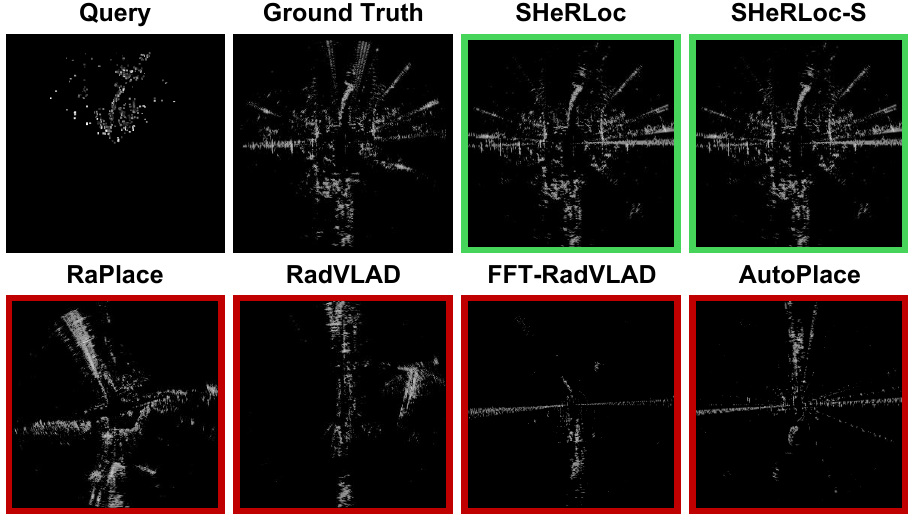}
    \caption{Qualitative results of \ac{SOTA} methods on a challenging query. Green indicates successful retrieval, and red indicates failure.}
    \label{fig: qualitative}
    \vspace{-5mm}
\end{figure}

\subsubsection{Heterogeneous Radar Multi-session PR}
\begin{figure}[t!]
    \centering
    \includegraphics[trim= 0cm 0.6cm 0cm 0cm, clip, width=1.0\columnwidth]{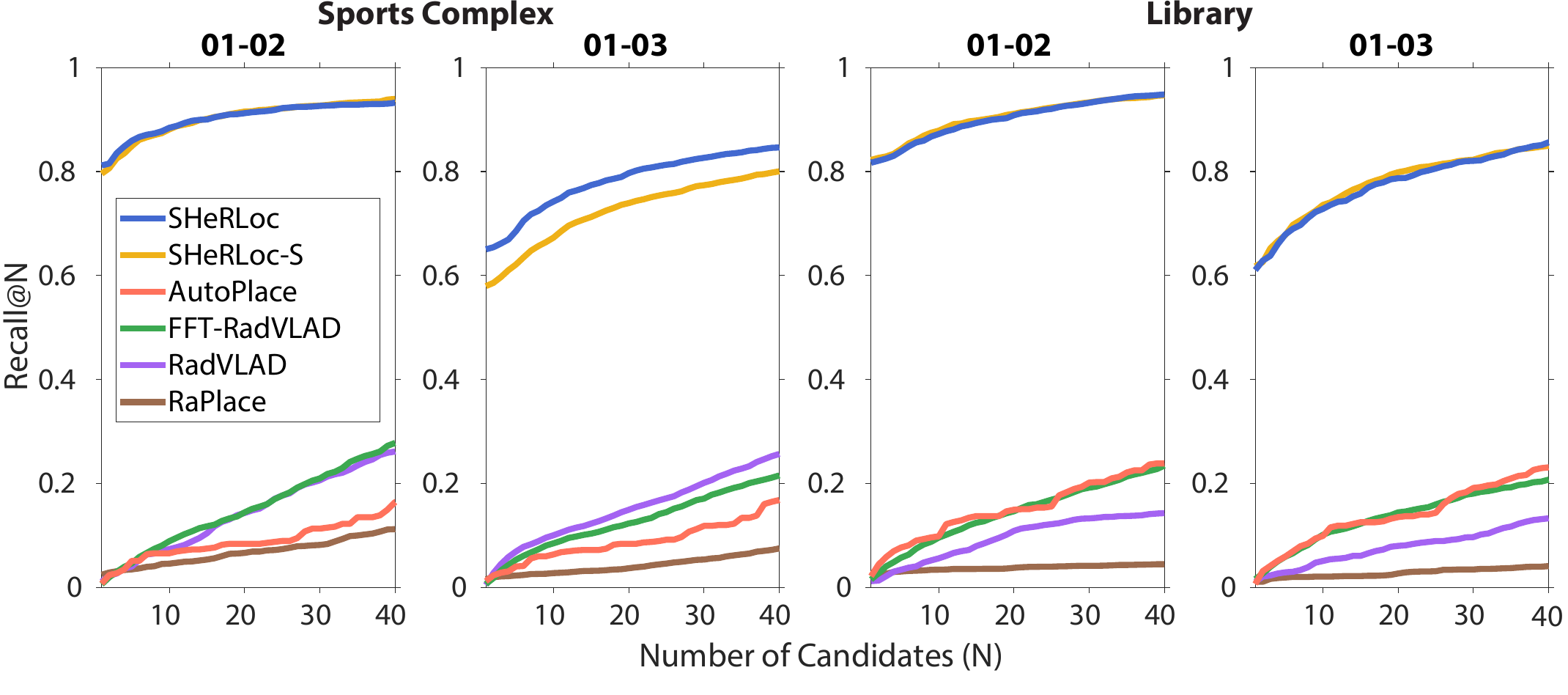}
    \caption{R@N curves for multi-session PR with heterogeneous radar. SHeRLoc outperforms all methods when retrieving 40 candidates.}
    \label{fig: R@N}
    \vspace{-4mm}
\end{figure}

For multi-session evaluation, spinning radar data from sequence \texttt{01} were used as the database, and 4D radar data from sequences \texttt{02} and \texttt{03} were used as queries. Quantitative results are presented in \tabref{tab: Multi}, with qualitative analysis shown in \figref{fig: qualitative} and recall@N curves depicted in \figref{fig: R@N}.
We have demonstrated the necessity of heterogeneous radar PR models beyond homogeneous radar systems by showing that existing \ac{SOTA} methods fail to address the inherent challenges of heterogeneous radar.

\subsubsection{Homogeneous Radar Multi-session PR}

\begin{figure}[t!]
    \centering
    \includegraphics[width=1.0\columnwidth]{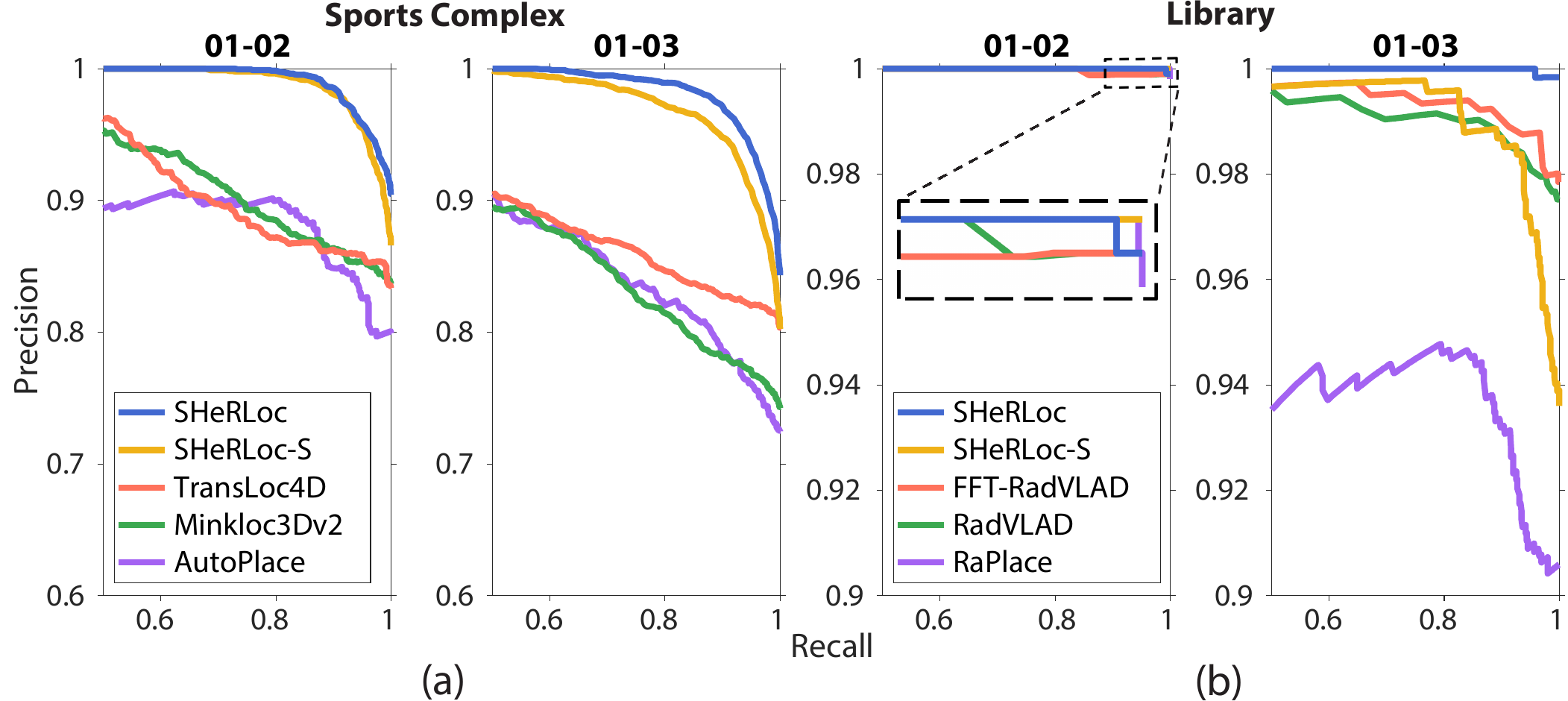}
    \caption{Precision-Recall curve for homogeneous radar multi-session PR: (a) 4D radar-based PR results on the \texttt{Sports Complex}, (b) Spinning radar-based PR results on the \texttt{Library}.}
    \label{fig: pr curve}
    \vspace{-6mm}
\end{figure}

SHeRLoc was compared with Autoplace~\cite{cai2022autoplace}, MinkLoc3Dv2~\cite{komorowski2022improving}, and TransLoc4D~\cite{peng2024transloc4d} for 4D radar PR, 
and with RaPlace~\cite{jang2023raplace}, RadVLAD, and FFT-RadVLAD~\cite{gadd2024open} for spinning radar PR. The results are reported in \tabref{tab: Multi} and \figref{fig: pr curve}, confirming SHeRLoc's superior performance despite not being designed for homogeneous radar. For 4D radar PR, it demonstrates top performance across nearly all sequences. For spinning radar PR, SHeRLoc achieves an AR@1 of 0.993, surpassing all \ac{SOTA} methods. Furthermore, PR curves indicate that SHeRLoc achieves high precision and recall for both radar types under diverse conditions, demonstrating robust performance for both homogeneous and heterogeneous radar.

\begin{table}[!t]
\caption{Performance Comparison for Multi-session Place Recognition with Heterogeneous Radar and Homogeneous Radar}
\label{tab: Multi}
\centering
\resizebox{\columnwidth}{!}{
\begin{tabular}{ll|cccccccc|c} 
\toprule

&\multirow{3}{*}{\begin{tabular}[c]{@{}c@{}}Methods\end{tabular}}&
  \multicolumn{4}{c}{\texttt{Sports Complex}} &
  \multicolumn{4}{c|}{\texttt{Library}}
 \\
& &
\multicolumn{2}{c}{01 - 02}&
\multicolumn{2}{c}{01 - 03}&
\multicolumn{2}{c}{01 - 02}&
\multicolumn{2}{c|}{01 - 03}&
AR@1
  \\ \cline{3-10} \rule{0pt}{2.5ex}

 & &
  R@1 &
  R@1\% &
  R@1 &
  R@1\% &
  R@1 &
  R@1\% &
  R@1 &
  R@1\% 
 \\ \midrule

\multirow{7}{*}{\rotatebox{90}{Heterogeneous}}

& RaPlace~\cite{jang2023raplace}
& 0.024 & 0.056 & 0.016 & 0.057
& 0.021 & 0.044 & 0.011 & 0.023  
& 0.018\\ 

& RadVLAD~\cite{gadd2024open}
& 0.011 & 0.118 & 0.005 & 0.214
& 0.012 & 0.138 & 0.011 & 0.073
& 0.010\\ 

& FFT-RadVLAD~\cite{gadd2024open}
& 0.006 & 0.121 & 0.007 & 0.182
& 0.015 & 0.217 & 0.016 & 0.136 
& 0.011\\ 

& Autoplace~\cite{cai2022autoplace}
& 0.007 & 0.071 & 0.012 & 0.070
& 0.022 & 0.129 & 0.007 & 0.119 
& 0.012\\

\cmidrule{2-11}
&    
\cellcolor[HTML]{f3f7fc}SHeRLoc-S
& \cellcolor[HTML]{ecf8f1}\underline{0.796}
& \cellcolor[HTML]{ecf8f1}\underline{0.890}
& \cellcolor[HTML]{ecf8f1}\underline{0.580}
& \cellcolor[HTML]{ecf8f1}\underline{0.696}
& \cellcolor[HTML]{def3e6}\textbf{0.822}
& \cellcolor[HTML]{def3e6}\textbf{0.892}
& \cellcolor[HTML]{def3e6}\textbf{0.618}
& \cellcolor[HTML]{def3e6}\textbf{0.757} 
& \cellcolor[HTML]{ecf8f1}\underline{0.704}
\\

& 
  \cellcolor[HTML]{f3f7fc}SHeRLoc
& \cellcolor[HTML]{def3e6}\textbf{0.812}
& \cellcolor[HTML]{def3e6}\textbf{0.893}
& \cellcolor[HTML]{def3e6}\textbf{0.650}
& \cellcolor[HTML]{def3e6}\textbf{0.759}
& \cellcolor[HTML]{ecf8f1}\underline{0.817}
& \cellcolor[HTML]{ecf8f1}\underline{0.887}
& \cellcolor[HTML]{ecf8f1}\underline{0.610}
& \cellcolor[HTML]{ecf8f1}\underline{0.743} & \cellcolor[HTML]{def3e6}\textbf{0.722}\\

\midrule

\multirow{6}{*}{\rotatebox{90}{4D}}
&
Autoplace~\cite{cai2022autoplace}
& 0.799
& 0.967
& 0.725
& 0.945
& 0.812
& 0.986
& 0.619
& 0.901 
& 0.738\\ 

&
MinkLoc3Dv2~\cite{komorowski2022improving}
& 0.837
& \cellcolor[HTML]{def3e6}\textbf{0.982}
& 0.743
& \cellcolor[HTML]{def3e6}\textbf{0.977}
& 0.725
& 0.981
& 0.619
& \cellcolor[HTML]{ecf8f1}\underline{0.963}
& 0.735\\ 

&
TransLoc4D~\cite{peng2024transloc4d}
& 0.833
& \cellcolor[HTML]{ecf8f1}\underline{0.970}
& \cellcolor[HTML]{ecf8f1}\underline{0.804}
& \cellcolor[HTML]{ecf8f1}\underline{0.976}
& 0.801
& \cellcolor[HTML]{ecf8f1}\underline{0.991}
& 0.676
& 0.940 
& 0.779\\

\cmidrule{2-11}
&
  \cellcolor[HTML]{f3f7fc}SHeRLoc-S
& \cellcolor[HTML]{ecf8f1}\underline{0.866}
& 0.945
& 0.803
& 0.925
& \cellcolor[HTML]{ecf8f1}\underline{0.914}
& 0.986
& \cellcolor[HTML]{ecf8f1}\underline{0.908}
& \cellcolor[HTML]{def3e6}\textbf{0.988} 
& \cellcolor[HTML]{ecf8f1}\underline{0.873}
\\

& \cellcolor[HTML]{f3f7fc}SHeRLoc
& \cellcolor[HTML]{def3e6}\textbf{0.904}
& 0.961
& \cellcolor[HTML]{def3e6}\textbf{0.843}
& 0.949
& \cellcolor[HTML]{def3e6}\textbf{0.950}
& \cellcolor[HTML]{def3e6}\textbf{0.995}
& \cellcolor[HTML]{def3e6}\textbf{0.923}
& \cellcolor[HTML]{def3e6}\textbf{0.988}
& \cellcolor[HTML]{def3e6}\textbf{0.905}
\\
\midrule

\multirow{6}{*}{\rotatebox{90}{Spinning}}
&
RaPlace~\cite{jang2023raplace}
& 0.943 & 0.962 & 0.927 & 0.989
& 0.998 & \cellcolor[HTML]{def3e6}\textbf{1.000} & 0.906 & 0.955 
& 0.944\\ 

&
RadVLAD~\cite{gadd2024open}
& \cellcolor[HTML]{def3e6}\textbf{0.996} & \cellcolor[HTML]{def3e6}\textbf{1.000}
& \cellcolor[HTML]{ecf8f1}\underline{0.981} & \cellcolor[HTML]{def3e6}\textbf{1.000}
& \cellcolor[HTML]{ecf8f1}\underline{0.999} & \cellcolor[HTML]{def3e6}\textbf{1.000}
& 0.975
& \cellcolor[HTML]{def3e6}\textbf{1.000} 
& 0.988\\

&
FFT-RadVLAD~\cite{gadd2024open}
& \cellcolor[HTML]{def3e6}\textbf{0.996}
& \cellcolor[HTML]{def3e6}\textbf{1.000}
& \cellcolor[HTML]{def3e6}\textbf{0.985}
& \cellcolor[HTML]{def3e6}\textbf{1.000} 
& \cellcolor[HTML]{ecf8f1}\underline{0.999} & \cellcolor[HTML]{def3e6}\textbf{1.000}
& \cellcolor[HTML]{ecf8f1}\underline{0.979} & \cellcolor[HTML]{def3e6}\textbf{1.000} 
& \cellcolor[HTML]{ecf8f1}\underline{0.990}
\\ 


\cmidrule{2-11}

& \cellcolor[HTML]{f3f7fc}SHeRLoc-S
& 0.963
& \cellcolor[HTML]{ecf8f1}\underline{0.993}
& 0.962
& 0.981
& \cellcolor[HTML]{def3e6}\textbf{1.000}
& \cellcolor[HTML]{def3e6}\textbf{1.000}
& 0.936
& \cellcolor[HTML]{ecf8f1}\underline{0.997}
& 0.965
\\

&
  \cellcolor[HTML]{f3f7fc}SHeRLoc
& \cellcolor[HTML]{ecf8f1}\underline{0.993}
& \cellcolor[HTML]{def3e6}\textbf{1.000}
& \cellcolor[HTML]{ecf8f1}\underline{0.981}
& \cellcolor[HTML]{ecf8f1}\underline{0.999}
& \cellcolor[HTML]{ecf8f1}\underline{0.999}
& \cellcolor[HTML]{def3e6}\textbf{1.000}
& \cellcolor[HTML]{def3e6}\textbf{0.998}
& \cellcolor[HTML]{def3e6}\textbf{1.000}
& \cellcolor[HTML]{def3e6}\textbf{0.993}\\

\bottomrule
\end{tabular}
}
\vspace{-4mm}
\end{table}


\subsection{Robustness against Random Rotation}
\label{sec: Robustness}
In real-world scenarios, random yaw angle rotations between the database and query are common, making rotation robustness crucial. Therefore, we conducted experiments with random yaw angle rotations on database scans within a $360^\circ$ range. Through polar domain multi-view generation and the rotation-invariant design of the HOLMES feature aggregation network, SHeRLoc demonstrates superior rotation robustness compared to other methods. As shown in \tabref{tab: multi-view}, comparing AR@1 with \ac{SOTA} methods on \texttt{Sports Complex}, \texttt{Library}, and \texttt{River Island}, both SHeRLoc and SHeRLoc-S consistently achieve the best performance. This robust performance, despite angle variations, demonstrates that a limited number of multi-views is sufficient for achieving robust rotation invariance.

\begin{table}[t!]
\caption{Performance Comparison under Random Rotation}
\label{tab: multi-view}
\centering
\resizebox{\columnwidth}{!}{%
\begin{tabular}{l|ccccccccc} 
\toprule

  \multirow{2}{*}{Methods} &
  \multicolumn{2}{c}{\texttt{Sports Complex}} &
  \multicolumn{2}{c}{\texttt{Library}} &
  \multicolumn{2}{c}{\texttt{River Island}} \\ \cline{2-7} \rule{0pt}{2.5ex}

 &
  AR@1 &
  AR@1\% &
  AR@1 &
  AR@1\% &
  AR@1 &
  AR@1\% \\ \midrule




Radar SC~\cite{kim2020mulran}
& 0.045
& 0.087
& 0.041
& 0.063
& 0.007
& 0.021 \\ 
RaPlace~\cite{jang2023raplace}
& 0.030 
& 0.204 
& 0.130 
& 0.229 
& 0.002
& 0.191 \\ 
RadVLAD~\cite{gadd2024open}
& 0.011 
& 0.318
& 0.012
& 0.249 
& 0.016
& 0.402 
\\ 
FFT-RadVLAD~\cite{gadd2024open}
& 0.019 
& 0.310
& 0.012 
& 0.234 
& 0.019
& 0.429
\\ 
Autoplace~\cite{cai2022autoplace}
& 0.026 
& 0.161 
& 0.020 
& 0.127 
& 0.001
& 0.122
\\ 

\midrule



\cellcolor[HTML]{f3f7fc}SHeRLoc-S
& \cellcolor[HTML]{ecf8f1}\underline{0.905}
& \cellcolor[HTML]{ecf8f1}\underline{0.959}
& \cellcolor[HTML]{ecf8f1}\underline{0.828}
& \cellcolor[HTML]{ecf8f1}\underline{0.893}
& \cellcolor[HTML]{ecf8f1}\underline{0.710}
& \cellcolor[HTML]{ecf8f1}\underline{0.880} \\



\cellcolor[HTML]{f3f7fc}SHeRLoc
& \cellcolor[HTML]{def3e6}\textbf{0.918}
& \cellcolor[HTML]{def3e6}\textbf{0.964}
& \cellcolor[HTML]{def3e6}\textbf{0.858}
& \cellcolor[HTML]{def3e6}\textbf{0.920}
& \cellcolor[HTML]{def3e6}\textbf{0.733}
& \cellcolor[HTML]{def3e6}\textbf{0.904} \\

\bottomrule
\end{tabular}
}
\vspace{-2.5mm}
\end{table}



\begin{table}[t!]
\caption{Zero-shot Performance on Unseen Datasets (R@1)}
\label{tab: zeroshot}
\centering
\resizebox{\columnwidth}{!}{%
\begin{tabular}{l|cc|cc|c} 
\toprule

  \multirow{2}{*}{Methods} &
  \multicolumn{2}{c|}{\texttt{MulRan}} &
  \multicolumn{2}{c|}{\texttt{Oxford Radar}} &
  \multirow{2}{*}{AR@1} 
  
  \\  \rule{0pt}{2.5ex}

 &
  \texttt{DCC 01} &
  \texttt{KAIST 03}  &
  \texttt{\#1} to \texttt{\#3} &
  \texttt{\#2} to \texttt{\#3} \\
  \midrule

Radar SC~\cite{kim2020mulran}
& 0.683
& 0.896
& 0.655
& 0.348
& 0.646
\\ 
RaPlace~\cite{jang2023raplace}
& 0.641
& 0.932
& 0.666
& 0.437
& 0.669
 \\ 
RadVLAD~\cite{gadd2024open}
& 0.507
& 0.833
& 0.813
& 0.693
& 0.712
\\ 
FFT-RadVLAD~\cite{gadd2024open}
& 0.706
& 0.948
& 0.666
& 0.427
& 0.687
\\ 


\midrule

\cellcolor[HTML]{f3f7fc}SHeRLoc-S
& \cellcolor[HTML]{def3e6}\textbf{0.732}
& \cellcolor[HTML]{ecf8f1}\underline{0.961}
& \cellcolor[HTML]{ecf8f1}\underline{0.855}
& \cellcolor[HTML]{ecf8f1}\underline{0.748}
& \cellcolor[HTML]{ecf8f1}\underline{0.824}
 \\

\cellcolor[HTML]{f3f7fc}SHeRLoc
& \cellcolor[HTML]{ecf8f1}\underline{0.708}
& \cellcolor[HTML]{def3e6}\textbf{0.963}
& \cellcolor[HTML]{def3e6}\textbf{0.900}
& \cellcolor[HTML]{def3e6}\textbf{0.890}
& \cellcolor[HTML]{def3e6}\textbf{0.865}
 \\

\bottomrule
\end{tabular}
}
\vspace{-2mm}
\end{table}

\begin{table}[t!]
\caption{Performance Comparison for Place Recognition with Spinning Radar as Database and FMCW LiDAR as Query}
\label{tab: lidar}
\centering
\resizebox{\columnwidth}{!}{%
\begin{tabular}{l|ccccccccc} 
\toprule

  \multirow{2}{*}{Methods} &
  \multicolumn{2}{c}{\texttt{Sports Complex}} &
  \multicolumn{2}{c}{\texttt{Library}} &
  \multicolumn{2}{c}{\texttt{River Island}} \\ \cline{2-7} \rule{0pt}{2.5ex}

 &
  AR@1 &
  AR@1\% &
  AR@1 &
  AR@1\% &
  AR@1 &
  AR@1\% \\ \midrule




Radar SC~\cite{kim2020mulran}
& 0.141
& 0.188
& 0.046
& 0.060
& 0.033
& 0.050  \\ 

Autoplace~\cite{cai2022autoplace}
& 0.017
& 0.105
& 0.014
& 0.059
& 0.004
& 0.041  \\ 

Radar-to-LiDAR~\cite{yin2021radar}
& \cellcolor[HTML]{ecf8f1}\underline{0.656}
& {0.832}
& {0.468}
& {0.769}
& {0.229}
& {0.530}  \\

\midrule










\cellcolor[HTML]{f3f7fc}SHeRLoc-S
& \cellcolor[HTML]{def3e6}\textbf{0.948}
& \cellcolor[HTML]{ecf8f1}\underline{0.971}
& \cellcolor[HTML]{ecf8f1}\underline{0.895}
& \cellcolor[HTML]{def3e6}\textbf{0.945}
& \cellcolor[HTML]{ecf8f1}\underline{0.932}
& \cellcolor[HTML]{ecf8f1}\underline{0.983} \\










\cellcolor[HTML]{f3f7fc}SHeRLoc
& \cellcolor[HTML]{def3e6}\textbf{0.948}
& \cellcolor[HTML]{def3e6}\textbf{0.977}
& \cellcolor[HTML]{def3e6}\textbf{0.899}
& \cellcolor[HTML]{ecf8f1}\underline{0.931}
& \cellcolor[HTML]{def3e6}\textbf{0.952}
& \cellcolor[HTML]{def3e6}\textbf{0.987} \\

\bottomrule
\end{tabular}
}
\vspace{-3mm}
\end{table}



\subsection{Zero-shot Generalization Performance Evaluation}
We evaluated SHeRLoc’s zero-shot generalization on unseen datasets, including MulRan~\cite{kim2020mulran} and the Oxford Radar RobotCar~\cite{barnes2020oxford}. 
While HeRCULES~\cite{kim2025hercules} employs the Navtech RAS6 spinning radar, MulRan uses the earlier Navtech CIR204-H, and Oxford Radar RobotCar uses the Navtech CTS350-X. 
These hardware differences induce substantial domain shifts in terms of noise characteristics, resolution, and detection range.
For MulRan, we conducted single-session PR on the \texttt{DCC 01} and \texttt{KAIST 03} sequences. 
For Oxford Radar RobotCar, we performed multi-session PR using \texttt{2019-01-18 (\#3)} as the database and \texttt{2019-01-10 (\#1)} and \texttt{2019-01-16 (\#2)} as queries. 
As shown in \tabref{tab: zeroshot}, SHeRLoc achieves strong recall even under challenging zero-shot settings with
domain shifts.

\subsection{Heterogeneous Range Sensor Place Recognition}
\begin{figure}[t!]
    \centering
    \includegraphics[width=\columnwidth]{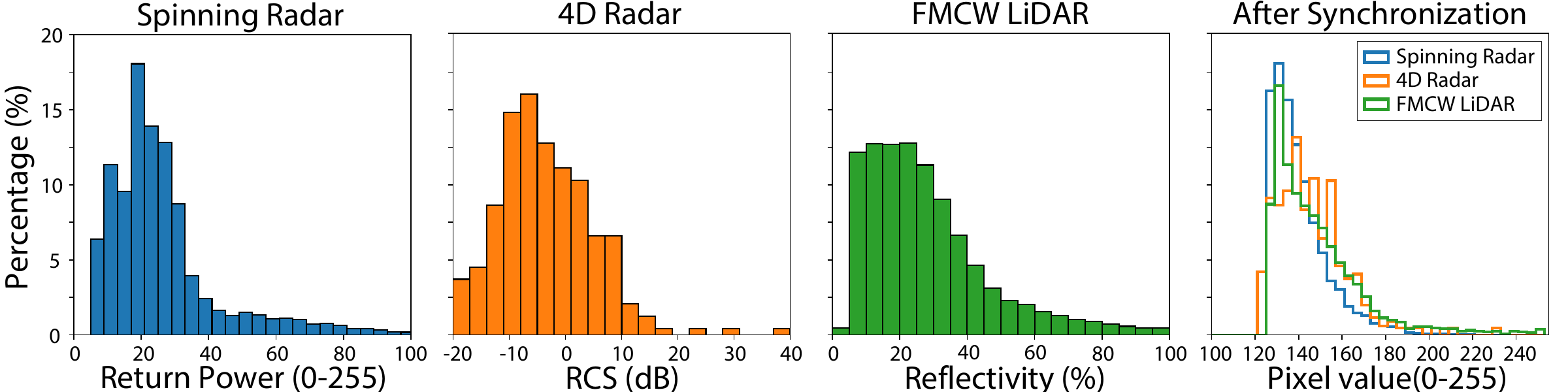}
    \caption{Data distribution of each sensor and the synchronized data distribution after SHeRLoc's \ac{RCS} polar domain matching.}
    \label{fig: sync}
    \vspace{-3mm}
\end{figure}

We evaluated heterogeneous range sensor PR performance using \ac{FMCW} LiDAR queries and a spinning radar database, comparing it with Radar Scan Context~\cite{kim2020mulran}, Autoplace~\cite{cai2022autoplace}, and Radar-to-LiDAR~\cite{yin2021radar}. We utilized radial velocity in a manner analogous to 4D radar, and replaced \ac{RCS} with reflectivity \(\rho\), which satisfies the following LiDAR equation:
\vspace{-1mm}
\begin{equation}\small
I \propto \frac{P_t D_r^2 \eta^2 \rho \cos \alpha_i}{4 R^2},
\label{eqn:lidar_intensity}
\end{equation}
\normalsize
where \(I\) is the LiDAR intensity, \(P_t\) is the transmitted power, \(D_r\) is the receiver aperture diameter, \(\eta\) is the system transmission factor, \(\alpha_i\) is the angle of incidence.
Our preprocessing aligns the distributions of \ac{FMCW} LiDAR and spinning radar, as shown in \figref{fig: sync}, and \tabref{tab: lidar} confirms that SHeRLoc remains robust across heterogeneous range sensors.


\begin{table}[t]
\caption{Ablation Study on SHeRLoc Components to Evaluate Individual Contributions}
\label{tab: component}
\centering
\resizebox{\columnwidth}{!}{%
\begin{tabular}{ccccc|cccccccc} 
\toprule

\multirow{3}{*}{B} & 
\multirow{3}{*}{S} & 
\multirow{3}{*}{P} & 
\multirow{3}{*}{R} & 
\multirow{3}{*}{A} & 
\multicolumn{4}{c}{\texttt{Sports Complex}} &
\multicolumn{4}{c}{\texttt{Library}}
 \\
 
& & & & &
\multicolumn{2}{c}{01 - 02}&
\multicolumn{2}{c}{01 - 03}&
\multicolumn{2}{c}{01 - 02}&
\multicolumn{2}{c}{01 - 03}
  \\ \cline{6-13} \rule{0pt}{2.5ex}

 & & & & &
  R@1 &
  R@1\% &
  R@1 &
  R@1\% &
  R@1 &
  R@1\% &
  R@1 &
  R@1\% 
 \\ \midrule

\checkmark  &
$\times$  &
$\times$  &
$\times$  &
$\times$  
& 0.009
& 0.015
& 0.009
& 0.013
& 0.019
& 0.032
& 0.016
& 0.035 \\

\checkmark  &
  \checkmark &
$\times$  &
$\times$  &
$\times$  
& 0.013
& 0.072
& 0.009
& 0.032
& 0.011
& 0.022
& 0.008
& 0.031 \\

\checkmark  &
  \checkmark  &
   \checkmark &
$\times$  &
$\times$  & 0.659
& 0.792
& 0.516
& 0.633
& 0.657
& 0.749
& 0.497
& 0.665 \\

\checkmark  &
  \checkmark &
  \checkmark &
  \checkmark &
$\times$  & 0.737
& 0.840
& 0.529
& 0.625
& 0.751
& 0.836
& 0.514
& 0.679 \\ 

$\times$  &
\checkmark  &
\checkmark  &
\checkmark  &
\checkmark  
& 0.558
& 0.748
& 0.295
& 0.426
& 0.511
& 0.631
& 0.199
& 0.350 \\ 

\midrule

  \cellcolor[HTML]{f3f7fc}\checkmark  &  \cellcolor[HTML]{f3f7fc}\checkmark  &
  \cellcolor[HTML]{f3f7fc}\checkmark  &
  \cellcolor[HTML]{f3f7fc}\checkmark  &
  \cellcolor[HTML]{f3f7fc}\checkmark  
& \cellcolor[HTML]{def3e6}\textbf{0.812}
& \cellcolor[HTML]{def3e6}\textbf{0.893}
& \cellcolor[HTML]{def3e6}\textbf{0.650}
& \cellcolor[HTML]{def3e6}\textbf{0.759}
& \cellcolor[HTML]{def3e6}\textbf{0.817}
& \cellcolor[HTML]{def3e6}\textbf{0.887}
& \cellcolor[HTML]{def3e6}\textbf{0.610}
& \cellcolor[HTML]{def3e6}\textbf{0.743} \\

\bottomrule
\end{tabular}
}
\vspace{-2mm}

\end{table}


\begin{table}[t!]
\caption{Ablation Study on Feature Aggregation Method}
\label{tab:aggregation}
\centering
\resizebox{\columnwidth}{!}{%
\begin{tabular}{l|c|ccccccccccc} 
\toprule

  \multirow{3}{*}{Methods} &
  \multirow{3}{*}{Size} &
  \multicolumn{4}{c}{\texttt{Sports Complex}} &
  \multicolumn{4}{c}{\texttt{Library}}\\ 
  
 &
 &
  \multicolumn{2}{c}{01 - 02} &
  \multicolumn{2}{c}{01 - 03} &
  \multicolumn{2}{c}{01 - 02} &
  \multicolumn{2}{c}{01 - 03} \\ 
  \cline{3-10} \rule{0pt}{2.5ex}
  
 &
 &
  R@1 &
  R@1\% &
  R@1 &
  R@1\% &
  R@1 &
  R@1\% &
  R@1 &
  R@1\%\\ \midrule




NetVLAD~\cite{arandjelovic2016netvlad}
& 32768 
& 0.378
& 0.598
& 0.151
& 0.254
& 0.302
& 0.474
& 0.102
& 0.218 \\

MAC~\cite{tolias2015particular}
& 512
& 0.700
& 0.846
& 0.404
& 0.534
& 0.525
& 0.657
& 0.258
& 0.514 \\ 

SPoC~\cite{babenko2015aggregating}
& 512 
& 0.282
& 0.517
& 0.112
& 0.210
& 0.155
& 0.258
& 0.036
& 0.150\\ 

GeM~\cite{radenovic2018fine}
& 512 
& 0.486
& 0.717
& 0.234
& 0.384
& 0.324
& 0.475
& 0.091
& 0.197 \\ 

SALAD~\cite{izquierdo2024optimal}
& 8448 
& 0.371
& 0.611
& 0.170
& 0.295
& 0.330
& 0.497
& 0.094
& 0.235 \\

\midrule
\cellcolor[HTML]{f3f7fc}HOLMES-S*
& \cellcolor[HTML]{def3e6}\textbf{256}
& 0.769
& 0.882
& \cellcolor[HTML]{ecf8f1}\underline{0.616}
& \cellcolor[HTML]{ecf8f1}\underline{0.730}
& \cellcolor[HTML]{def3e6}\textbf{0.840}
& \cellcolor[HTML]{def3e6}\textbf{0.909}
& 0.587
& \cellcolor[HTML]{ecf8f1}\underline{0.756}
\\

\cellcolor[HTML]{f3f7fc}HOLMES-S
& \cellcolor[HTML]{def3e6}\textbf{256}
& \cellcolor[HTML]{ecf8f1}\underline{0.796}
& \cellcolor[HTML]{ecf8f1}\underline{0.890}
& 0.580
& 0.696
& \cellcolor[HTML]{ecf8f1}\underline{0.822}
& \cellcolor[HTML]{ecf8f1}\underline{0.892}
& \cellcolor[HTML]{def3e6}\textbf{0.618}
& \cellcolor[HTML]{def3e6}\textbf{0.757}  \\

\cellcolor[HTML]{f3f7fc}HOLMES
& \cellcolor[HTML]{ecf8f1}\underline{320}
& \cellcolor[HTML]{def3e6}\textbf{0.812}
& \cellcolor[HTML]{def3e6}\textbf{0.893}
& \cellcolor[HTML]{def3e6}\textbf{0.650}
& \cellcolor[HTML]{def3e6}\textbf{0.759}
& 0.817
& 0.887
& \cellcolor[HTML]{ecf8f1}\underline{0.610}
& 0.743 \\

\bottomrule
\multicolumn{8}{c}{HOLMES-S*: HOLMES-S without ghostbin and adaptive entropy regularization.}
\end{tabular}
}

\vspace{-3mm}
\end{table}


\subsection{Ablation Study}
\subsubsection{BSPRA Components}
We compared SHeRLoc variants that incorporate the following modules: \textbf{B} (Backbone sharing), \textbf{S} (Scan aggregation), 
\textbf{P} (Polar projection), \textbf{R} (RCS matching), \textbf{A} (Adaptive margin loss). \tabref{tab: component} demonstrates how each component contributes to the performance. In the Cartesian domain, performance is significantly poor, and applying \textbf{S} alone fails to yield significant improvement. However, combining \textbf{S} with \textbf{P}, as in our approach, provides an average improvement of 57.2\% in recall@1, with \textbf{R} further enhancing performance by 5.05\%.
As illustrated in \figref{fig: sync}, although each sensor's raw data has different distributions, using \textbf{S}, \textbf{P}, and \textbf{R} leads to very similar data distributions.
Finally, \textbf{A} contributes an additional 8.95\% increase, resulting in an overall improvement of 71.2\%.
Moreover, \textbf{B} encourages learning a shared feature space, which enhances cross-modal generalization and avoids redundant parameterization.


\subsubsection{Feature Aggregation}
We evaluated HOLMES against NetVLAD~\cite{arandjelovic2016netvlad}, MAC~\cite{tolias2015particular}, SPoC~\cite{babenko2015aggregating}, GeM~\cite{radenovic2018fine}, and SALAD~\cite{izquierdo2024optimal}.
As shown in \tabref{tab:aggregation}, widely used methods like GeM pooling are vulnerable to noisy and sparse radar data. Additionally, NetVLAD and SALAD require large descriptor sizes, leading to longer processing times and higher storage demands. 
In contrast, HOLMES-S, with a compact 256-dimensional descriptor, achieves accurate PR performance through its radar-specific design, including ghostbin and adaptive entropy regularization. 
Moreover, HOLMES, with a descriptor size of $320$ dimensions, further improves performance by employing multi-scale descriptors. 
As a result, HOLMES outperforms other methods, while requiring less storage and generating each descriptor in just \unit{13}{ms}.

\section{Conclusion}
\label{sec:conclusion}

In this work, we introduce SHeRLoc, the first deep network for heterogeneous radar PR. SHeRLoc employs a novel preprocessing method to transform radar data into synchronized RCS polar \ac{BEV} representations and hierarchical optimal transport-based multi-scale descriptors. Our \ac{FOV}-aware data mining and adaptive margin-based triplet loss facilitate spatially and viewpoint-wise meaningful metric learning. Evaluations on diverse sequences demonstrate that SHeRLoc achieves robust performance in challenging environments and is applicable to both heterogeneous and homogeneous radar systems, as well as heterogeneous range sensor systems. 
As the first study on heterogeneous radar PR, SHeRLoc opens new opportunities for cross-modal place recognition, enabling efficient localization of vehicles with automotive 4D radars on pre-built spinning radar maps.


\balance
\footnotesize
\bibliographystyle{IEEEtranN} 
\bibliography{string-short,references}

\begin{thebibliography}{39}
\providecommand{\natexlab}[1]{#1}
\providecommand{\url}[1]{#1}
\csname url@samestyle\endcsname
\providecommand{\newblock}{\relax}
\providecommand{\bibinfo}[2]{#2}
\providecommand{\BIBentrySTDinterwordspacing}{\spaceskip=0pt\relax}
\providecommand{\BIBentryALTinterwordstretchfactor}{4}
\providecommand{\BIBentryALTinterwordspacing}{\spaceskip=\fontdimen2\font plus
\BIBentryALTinterwordstretchfactor\fontdimen3\font minus \fontdimen4\font\relax}
\providecommand{\BIBforeignlanguage}[2]{{%
\expandafter\ifx\csname l@#1\endcsname\relax
\typeout{** WARNING: IEEEtranN.bst: No hyphenation pattern has been}%
\typeout{** loaded for the language `#1'. Using the pattern for}%
\typeout{** the default language instead.}%
\else
\language=\csname l@#1\endcsname
\fi
#2}}
\providecommand{\BIBdecl}{\relax}
\BIBdecl

\bibitem[Arandjelovic et~al.(2016)Arandjelovic, Gronat, Torii, Pajdla, and Sivic]{arandjelovic2016netvlad}
R.~Arandjelovic, P.~Gronat, A.~Torii, T.~Pajdla, and J.~Sivic, ``Netvlad: Cnn architecture for weakly supervised place recognition,'' in \emph{Proc. {IEEE} Conf. on Comput. Vision and Pattern Recog.}, 2016, pp. 5297--5307.

\bibitem[Kim and Kim(2018)]{kim2018scan}
G.~Kim and A.~Kim, ``Scan context: Egocentric spatial descriptor for place recognition within 3d point cloud map,'' in \emph{Proc. {IEEE}/{RSJ} Intl. Conf. on Intell. Robots and Sys.}, 2018.

\bibitem[Barnes et~al.(2020)Barnes, Gadd, Murcutt, Newman, and Posner]{barnes2020oxford}
D.~Barnes, M.~Gadd, P.~Murcutt, P.~Newman, and I.~Posner, ``The oxford radar robotcar dataset: A radar extension to the oxford robotcar dataset,'' in \emph{Proc. {IEEE} Intl. Conf. on Robot. and Automat.}, 2020, pp. 6433--6438.

\bibitem[Kim et~al.(2020)Kim, Park, Cho, Jeong, and Kim]{kim2020mulran}
G.~Kim, Y.~S. Park, Y.~Cho, J.~Jeong, and A.~Kim, ``Mulran: Multimodal range dataset for urban place recognition,'' in \emph{Proc. {IEEE} Intl. Conf. on Robot. and Automat.}, 2020, pp. 6246--6253.

\bibitem[Barnes and Posner(2020)]{barnes2020under}
D.~Barnes and I.~Posner, ``Under the radar: Learning to predict robust keypoints for odometry estimation and metric localisation in radar,'' in \emph{Proc. {IEEE} Intl. Conf. on Robot. and Automat.}, 2020, pp. 9484--9490.

\bibitem[S{\u{a}}ftescu et~al.(2020)S{\u{a}}ftescu, Gadd, De~Martini, Barnes, and Newman]{suaftescu2020kidnapped}
{\c{S}}.~S{\u{a}}ftescu, M.~Gadd, D.~De~Martini, D.~Barnes, and P.~Newman, ``Kidnapped radar: Topological radar localisation using rotationally-invariant metric learning,'' in \emph{Proc. {IEEE} Intl. Conf. on Robot. and Automat.}, 2020, pp. 4358--4364.

\bibitem[Jang et~al.(2023)Jang, Jung, and Kim]{jang2023raplace}
H.~Jang, M.~Jung, and A.~Kim, ``Raplace: Place recognition for imaging radar using radon transform and mutable threshold,'' in \emph{Proc. {IEEE}/{RSJ} Intl. Conf. on Intell. Robots and Sys.}, 2023.

\bibitem[Gadd and Newman(2024)]{gadd2024open}
M.~Gadd and P.~Newman, ``Open-radvlad: Fast and robust radar place recognition,'' in \emph{2024 IEEE Radar Conference}, 2024, pp. 1--6.

\bibitem[Kim et~al.(2024)Kim, Choi, Choi, and Cho]{kim2024referee}
H.~Kim, B.~Choi, E.~Choi, and Y.~Cho, ``Referee: Radar-based lightweight and robust localization using feature and free space,'' \emph{{IEEE} Robot. and Automat. Lett.}, 2024.

\bibitem[Cai et~al.(2022)Cai, Wang, and Lu]{cai2022autoplace}
K.~Cai, B.~Wang, and C.~X. Lu, ``Autoplace: Robust place recognition with single-chip automotive radar,'' in \emph{Proc. {IEEE} Intl. Conf. on Robot. and Automat.}, 2022, pp. 2222--2228.

\bibitem[Meng et~al.(2024)Meng, Duan, He, Wang, Fan, and Zhang]{meng2024mmplace}
C.~Meng, Y.~Duan, C.~He, D.~Wang, X.~Fan, and Y.~Zhang, ``mmplace: Robust place recognition with intermediate frequency signal of low-cost single-chip millimeter wave radar,'' \emph{{IEEE} Robot. and Automat. Lett.}, 2024.

\bibitem[Herraez et~al.(2024)Herraez, Chang, Zeller, Wiesmann, Behley, Heidingsfeld, and Stachniss]{herraez2024spr}
D.~C. Herraez, L.~Chang, M.~Zeller, L.~Wiesmann, J.~Behley, M.~Heidingsfeld, and C.~Stachniss, ``Spr: Single-scan radar place recognition,'' \emph{{IEEE} Robot. and Automat. Lett.}, 2024.

\bibitem[Peng et~al.(2024)Peng, Li, Zhao, Zhang, Wu, Zheng, and Wang]{peng2024transloc4d}
G.~Peng, H.~Li, Y.~Zhao, J.~Zhang, Z.~Wu, P.~Zheng, and D.~Wang, ``Transloc4d: Transformer-based 4d radar place recognition,'' in \emph{Proc. {IEEE} Conf. on Comput. Vision and Pattern Recog.}, 2024, pp. 17\,595--17\,605.

\bibitem[Hilger et~al.(2025)Hilger, Kubelka, Adolfsson, Becker, Andreasson, and Lilienthal]{hilger2025introspective}
M.~Hilger, V.~Kubelka, D.~Adolfsson, R.~Becker, H.~Andreasson, and A.~J. Lilienthal, ``Introspective loop closure for slam with 4d imaging radar,'' \emph{arXiv preprint arXiv:2503.02383}, 2025.

\bibitem[Yao et~al.(2024)Yao, Li, Fu, and Pan]{yao2024monocular}
G.~Yao, X.~Li, L.~Fu, and Y.~Pan, ``Monocular visual place recognition in lidar maps via cross-modal state space model and multi-view matching,'' \emph{arXiv preprint arXiv:2410.06285}, 2024.

\bibitem[Nayak et~al.(2024)Nayak, Cattaneo, and Valada]{nayak2024ralf}
A.~Nayak, D.~Cattaneo, and A.~Valada, ``Ralf: Flow-based global and metric radar localization in lidar maps,'' in \emph{Proc. {IEEE} Intl. Conf. on Robot. and Automat.}, 2024, pp. 5097--5103.

\bibitem[Tang et~al.(2021)Tang, De~Martini, and Newman]{tang2021get}
T.~Y. Tang, D.~De~Martini, and P.~Newman, ``Get to the point: Learning lidar place recognition and metric localisation using overhead imagery,'' \emph{Proceedings of Robotics: Science and Systems, 2021}, 2021.

\bibitem[Cattaneo et~al.(2020)Cattaneo, Vaghi, Fontana, Ballardini, and Sorrenti]{cattaneo2020global}
D.~Cattaneo, M.~Vaghi, S.~Fontana, A.~L. Ballardini, and D.~G. Sorrenti, ``Global visual localization in lidar-maps through shared 2d-3d embedding space,'' in \emph{2020 IEEE International Conference on Robotics and Automation}.\hskip 1em plus 0.5em minus 0.4em\relax IEEE, 2020, pp. 4365--4371.

\bibitem[Xie et~al.(2024)]{xie2024modalink}
W.~Xie \emph{et~al.}, ``Modalink: Unifying modalities for efficient image-to-pointcloud place recognition,'' in \emph{Proc. {IEEE}/{RSJ} Intl. Conf. on Intell. Robots and Sys.}, 2024, pp. 3326--3333.

\bibitem[Zhou et~al.(2023)Zhou, Xu, Xiong, and Ma]{zhou2023lcpr}
Z.~Zhou, J.~Xu, G.~Xiong, and J.~Ma, ``Lcpr: A multi-scale attention-based lidar-camera fusion network for place recognition,'' \emph{IEEE Robotics and Automation Letters}, vol.~9, no.~2, pp. 1342--1349, 2023.

\bibitem[Jung et~al.(2025)Jung, Jung, Gil, and Kim]{jung2025helios}
M.~Jung, S.~Jung, H.~Gil, and A.~Kim, ``Helios: Heterogeneous lidar place recognition via overlap-based learning and local spherical transformer,'' \emph{arXiv preprint arXiv:2501.18943}, 2025.

\bibitem[Doer and Trommer(2020)]{doer2020ekf}
C.~Doer and G.~F. Trommer, ``An ekf based approach to radar inertial odometry,'' in \emph{2020 IEEE International Conference on Multisensor Fusion and Integration for Intelligent Systems}.\hskip 1em plus 0.5em minus 0.4em\relax IEEE, 2020, pp. 152--159.

\bibitem[Shi et~al.(2023)Shi, Shi, Yang, Yin, Lin, and Wang]{shi2023panovpr}
Z.~Shi, H.~Shi, K.~Yang, Z.~Yin, Y.~Lin, and K.~Wang, ``Panovpr: towards unified perspective-to-equirectangular visual place recognition via sliding windows across the panoramic view,'' in \emph{Proc. {IEEE} Intell. Transport. Sys. Conf.}, 2023, pp. 1333--1340.

\bibitem[Hao and He(2017)]{hao2017developing}
Z.-C. Hao and M.~He, ``Developing millimeter-wave planar antenna with a cosecant squared pattern,'' \emph{{IEEE} Trans. Antennas and Prop.}, vol.~65, no.~10, pp. 5565--5570, 2017.

\bibitem[He et~al.(2016)He, Zhang, Ren, and Sun]{he2016deep}
K.~He, X.~Zhang, S.~Ren, and J.~Sun, ``Deep residual learning for image recognition,'' in \emph{Proc. {IEEE} Conf. on Comput. Vision and Pattern Recog.}, 2016, pp. 770--778.

\bibitem[Izquierdo and Civera(2024)]{izquierdo2024optimal}
S.~Izquierdo and J.~Civera, ``Optimal transport aggregation for visual place recognition,'' in \emph{Proc. {IEEE} Conf. on Comput. Vision and Pattern Recog.}, 2024, pp. 17\,658--17\,668.

\bibitem[Zhong et~al.(2019)Zhong, Arandjelovi{\'c}, and Zisserman]{zhong2019ghostvlad}
Y.~Zhong, R.~Arandjelovi{\'c}, and A.~Zisserman, ``Ghostvlad for set-based face recognition,'' in \emph{Computer Vision--ACCV 2018: 14th Asian Conference on Computer Vision, Perth, Australia, December 2--6, 2018, Revised Selected Papers, Part II 14}.\hskip 1em plus 0.5em minus 0.4em\relax Springer, 2019, pp. 35--50.

\bibitem[Cuturi(2013)]{cuturi2013sinkhorn}
M.~Cuturi, ``Sinkhorn distances: Lightspeed computation of optimal transport,'' \emph{Advances in Neural Information Processing Sys. Conf.}, vol.~26, 2013.

\bibitem[Duhamel and Vetterli(1990)]{duhamel1990fast}
P.~Duhamel and M.~Vetterli, ``Fast fourier transforms: a tutorial review and a state of the art,'' \emph{Signal processing}, vol.~19, no.~4, pp. 259--299, 1990.

\bibitem[Xu et~al.(2023)Xu, Lu, Wu, Lu, Zhu, Liao, Xiong, and Wang]{xu2023ring++}
X.~Xu, S.~Lu, J.~Wu, H.~Lu, Q.~Zhu, Y.~Liao, R.~Xiong, and Y.~Wang, ``Ring++: Roto-translation invariant gram for global localization on a sparse scan map,'' \emph{IEEE Transactions on Robotics}, vol.~39, no.~6, pp. 4616--4635, 2023.

\bibitem[Cohen and Welling(2016)]{cohen2016group}
T.~Cohen and M.~Welling, ``Group equivariant convolutional networks,'' in \emph{International conference on machine learning}.\hskip 1em plus 0.5em minus 0.4em\relax PMLR, 2016, pp. 2990--2999.

\bibitem[Zhang et~al.(2023)]{zhang2023ntu4dradlm}
J.~Zhang \emph{et~al.}, ``Ntu4dradlm: 4d radar-centric multi-modal dataset for localization and mapping,'' in \emph{Proc. {IEEE} Intell. Transport. Sys. Conf.}, 2023, pp. 4291--4296.

\bibitem[Huai et~al.(2024)Huai, Wang, Zhuang, Chen, Li, and Han]{huai2024snail}
J.~Huai, B.~Wang, Y.~Zhuang, Y.~Chen, Q.~Li, and Y.~Han, ``Snail radar: A large-scale diverse benchmark for evaluating 4d-radar-based slam,'' \emph{The International Journal of Robotics Research}, p. 02783649251329048, 2024.

\bibitem[Kim et~al.(2025)]{kim2025hercules}
H.~Kim \emph{et~al.}, ``Hercules: Heterogeneous radar dataset in complex urban environment for multi-session radar slam,'' in \emph{Proc. of the IEEE International Conference on Robotics and Automation}, 2025.

\bibitem[Komorowski(2022)]{komorowski2022improving}
J.~Komorowski, ``Improving point cloud based place recognition with ranking-based loss and large batch training,'' in \emph{2022 26th international conference on pattern recognition}.\hskip 1em plus 0.5em minus 0.4em\relax IEEE, 2022, pp. 3699--3705.

\bibitem[Yin et~al.(2021)Yin, Xu, Wang, and Xiong]{yin2021radar}
H.~Yin, X.~Xu, Y.~Wang, and R.~Xiong, ``Radar-to-lidar: Heterogeneous place recognition via joint learning,'' \emph{Frontiers in Robotics and AI}, vol.~8, p. 661199, 2021.

\bibitem[Tolias et~al.(2015)Tolias, Sicre, and J{\'e}gou]{tolias2015particular}
G.~Tolias, R.~Sicre, and H.~J{\'e}gou, ``Particular object retrieval with integral max-pooling of cnn activations,'' \emph{arXiv preprint arXiv:1511.05879}, 2015.

\bibitem[Babenko and Lempitsky(2015)]{babenko2015aggregating}
A.~Babenko and V.~Lempitsky, ``Aggregating deep convolutional features for image retrieval,'' \emph{arXiv preprint arXiv:1510.07493}, 2015.

\bibitem[Radenovi{\'c} et~al.(2018)Radenovi{\'c}, Tolias, and Chum]{radenovic2018fine}
F.~Radenovi{\'c}, G.~Tolias, and O.~Chum, ``Fine-tuning cnn image retrieval with no human annotation,'' \emph{{IEEE} Trans. Pattern Analysis and Machine Intell.}, vol.~41, no.~7, pp. 1655--1668, 2018.

\end{thebibliography}

\end{document}